\documentclass[runningheads]{llncs}
\usepackage[sorting=none, backend=biber, url=false]{biblatex}
\usepackage[super]{nth}
\usepackage[toc,page]{appendix}
\usepackage{placeins}
\usepackage{float}
\usepackage{subcaption}
\usepackage{adjustbox}
\usepackage[T1]{fontenc}
\usepackage[english]{babel}
\usepackage{graphicx}

\usepackage{algpseudocode}
\usepackage{algorithm}

\usepackage{hyperref}
\usepackage{csquotes}
\usepackage{amsmath}
\usepackage[super]{nth}
\usepackage[dvipsnames]{xcolor}
\usepackage[x11names]{xcolor}

\usepackage{caption} 
\algnewcommand\algorithmicforeach{\textbf{for each}}
\algdef{S}[FOR]{ForEach}[1]{\algorithmicforeach\ #1\ \algorithmicdo}
\algdef{SE}[REPEATN]{RepeatN}{End}[1]{\algorithmicrepeat\ #1 \textbf{times}}{\algorithmicend}

\AtEveryBibitem{
	\clearfield{urlyear}
	\clearfield{urlmonth}
}

\authorrunning{Deproost et al.}
\titlerunning{Critic-driven Voronoi State Partitioning}

\begin{document}
\title{Critic-Driven Voronoi-Quantization for Distilling Deep RL Policies to Explainable Models}
%
%
\author{ Senne Deproost\inst{1, 2}\orcidID{0009-0009-4757-0290}\and \\
	Denis Steckelmacher \inst{1, 2}\orcidID{0000-0003-1521-8494}\and \\
	Ann Now\'{e}\inst{1, 2}\orcidID{0000-0001-6346-4564}}

\institute{Vrije Universiteit Brussel, Pleinlaan 2, 1050 Brussels, Belgium \and
	BP\&M, Flanders Make@VUB, Pleinlaan 2, 1050 Brussels, Belgium}
\maketitle              
\begin{abstract}
Despite many successful attempts at explaining Deep Reinforcement Learning policies using distillation, it remains difficult to balance the performance-interpretability trade-off and select a fitting surrogate model. 
In addition to this, traditional distillation only minimizes the distance between the behavior of the original and the surrogate policy while other RL-specific components such as action value are disregarded. 
To solve this, we introduce a new model-agnostic method called Critic-Driven Voronoi State Partitioning, which partitions a black box control policy into regions where a simple class of model can be optimized using gradient descent. By exploiting the critic value network of the original policy, we iteratively introduce new subpolicies in regions with insufficient value, standing in for a measure of policy complexity. The partitioning, a Voronoi quantizer, uses nearest neighbor lookups to assign a linear function to each point in the state space resulting in a cell-like diagram. We validate our approach on several well known benchmarks and proof that this distillation approaches the original policy using a reasonable sized set of linear functions.

\keywords{Reinforcement learning  \and State partitioning}
\end{abstract}

\section{Introduction}
Deep Reinforcement Learning (DRL) allows to learn high-quality controllers, directly from experience, using only a reward function \cite{mnih_etal_2015_HumanlevelControla, lillicrap_etal_2015_ContinuousControla}. DRL is particularly useful for producing controllers for systems that are highly stochastic, poorly-understood, difficult to model, involving people, or in which planning is challenging for any other reason \cite{padakandla_2022_SurveyReinforcement, recht_2019_TourReinforcement}.

However, state-of-the-art DRL methods are built on deep neural networks, black boxes widely considered to be unexplainable \cite{miller_2019_ExplanationArtificial}. This makes both local and global explanations difficult to produce. Local explanations aim at motivating why the controller choose this particular action in this particular state. Global explanations aim at summarizing the entire policy in a human-friendly way, such as sets of rules \cite{hein_etal_2018_GeneratingInterpretable} or source code \cite{verma_etal_2019_ImitationProjectedProgrammatic}.

The lack of explanations of DRL policies discourages users to deploy a learned DRL policy into production since no guarantees can be provided on stability, robustness, etc. Additionally, the deployment of deep neural networks is only feasible on hardware that supports them, which may not be the case of small microcontrollers in fast-sampled systems.

Explainable Reinforcement Learning (XRL) studies techniques that combine Reinforcement Learning with explainable representations \cite{bekkemoen_2023_ExplainableReinforcement}. Two methods prevail: using an explainable Machine Learning model in the RL agent (instead of a neural network), or training the DRL agent as usual, but, as a post-training step, distilling its policy into an explainable representation, also called surrogate model \cite{frosst_hinton_2017_DistillingNeural, rusu_etal_2015_PolicyDistillation}.

Common surrogate models, often hierarchical or rule based, are known to grow to large sizes, lowering interpretability. Besides, most distillation techniques rely solely on minimizing the distance between the behavior of both the surrogate and the original policy. In a Reinforcement Learning setting, more elements, such as the notion of value of an action, can be exploited, to improve the distillation process and to inform the user on RL-specific conditions.

We argue that the complexity of a learned policy is difficult to estimate a priori, making decisions on the class of surrogate not straightforward. In addition, a policy could require simple functions in large sections of its operation space while only needing more complexity elsewhere. A single model of fixed complexity could have over- and under-capacity in these regions, so one would rather opt for a more complex and capable model for the space as a whole. 

In this paper, we propose an offline distillation algorithm that partitions the state space into regions in which simple subpolicies can approximate a dataset of state-action pairs produced by the original (deep, black-box) policy. This is accomplished by alternating between training a current collection of subpolicies on their assigned partition of the data, and introducing new subpolicies and partitions in regions where they are needed. By exploiting the value function learned by the RL agent, subpolicies can be evaluated not only in how close they match the black-box policy, but also "how much worse" the actions they produce are.

The result is a partitioning algorithm that decides points of interest to put new subpolicies on with the decision boundary being made using the nearest neighbor or Voronoi quantization lookup.

Our empirical evaluation confirms that our method is able to distill high-quality DRL policies into an explainable representation, while preserving most of the quality of the policy. We also evaluate the impact of the hyper-parameters of our method on the tradeoff between explainability (small, simple model) and accuracy (good imitation of the DRL policy).

\section{Background}
\subsection{Deep Reinforcement Learning}
Reinforcement Learning (RL) is a machine learning technique for solving sequential decision problems \cite{sutton_barto_2014_ReinforcementLearning}. Its mathematical basis is the Markov Decision Process, defined by a tuple $ \left< S, A, P_a, R_a \right> $ \cite{bellman_1957_MarkovianDecision}. It consists of a state space $S$ and action space $A$ where the transition probability from $s$ to $s'$ by taking an action $a$ is given by $P_a(s, s')$. Additionally, an associate signal $r$ from the reward function $R_a(s, s')$ encodes the quality of the selected actions in accordance to the control objective. The control policy $\pi(a_t | s_t)$ iteratively interacts with the environment at every timestep $t \in \left[0, t_{max} \right]$. Over time, the algorithm seeks to find an optimal control policy that maximizes the sum of discounted rewards $R(\tau) = \sum \gamma^t r_{t + 1}$ with trajectory of states $\tau$ and $\gamma \in \left[0, 1\right[$ as the discount factor.

The behavior the policy $\pi$ can be optimized in a policy-driven manner \cite{kakade_2001_NaturalPolicy}, learning the mapping from state to action directly, value-based, where a value function estimates the expected return of an action \cite{watkins_dayan_1992_Qlearning}, or actor-critic, a combination of both \cite{konda_tsitsiklis_1999_ActorCriticAlgorithms}. For value estimation, a function $Q(s, a)$ is learned during training. It evaluates the expected sum of rewards obtained when executing an action $a$ in state $s$, and then following some policy $\pi$: $Q_{\pi}(s, a) \dot{=} E_{\pi}\left[ \sum_t \gamma^t r_t \right]$. The optimal Q-function $Q^*$ evaluates the optimal policy for the task. Usually, DRL algorithms learn only a good policy (not optimal), and the associated Q-function.

Current Reinforcement Learning algorithms, such as the Soft Actor-Critic \cite{haarnoja_etal_2018_SoftActorcritic} or the Proximal Policy Optimization \cite{schulman_etal_2017_ProximalPolicy}, rely on neural networks, especially in environments with continuous (real-valued) states and actions, as most modern environments are. This use of a model, widely recognized as a black box, is a major obstacle in user adoption since understanding the learned policy, or proving properties on it, is impossible.

\subsection{Explainable Reinforcement Learning}

Explainable Reinforcement Learning (XRL) aims and producing human-intelligible insights at all stages of training \cite{bekkemoen_2023_ExplainableReinforcement}. A most comprehensive taxonomy is to categorize techniques into three approaches: RL policies that are interpretable by design, methods that adapt the RL algorithm to make it explainable and approaches that gain insights after a black box policy has been trained. 

The first category is straightforward as it uses interpretable models as the policy from the beginning. A simple example is Tabular Q-Learning \cite{watkins_dayan_1992_Qlearning}, that, with a small amount of states and actions, allows a user to understand why an action is preferred in a state simply by looking at the table of learned Q-Values. Other methods include the use of decision trees \cite{dasgupta_etal_2015_PolicyTree}, or rule-based policies \cite{hein_etal_2018_GeneratingInterpretable}.

The second category extends the RL algorithm with an explanation generating component, providing insights during training as well. Here, more complex questions can be answered such as counterfactual (why is a certain behavior \textit{not} chosen) \cite{rupprecht_etal_2020_FindingVisualizing} and reward-driven explanations \cite{erwig_etal_2018_ExplainingDeep}. Using this setting, the user can intervene in the training when the agent is accomplishing the maximization of its return while not meeting certain criteria. If needed, the reward function can be reformulated or some safety measures can be enforced upon a new iteration of training.

A last approach is to retrieve insights in a trained policy without changing it. Here, the user can generate a (new) scenario for the policy and examine its the decisions it would take \cite{Mitsopoulos2021TowardAP, 10.1609/aaai.v39i15.33733}. It also encompasses techniques that consider a surrogate representation of the setting to be decomposed or transferred into a surrogate \cite{rusu_etal_2015_PolicyDistillation}. 

\subsection{Voronoi quantization}

\begin{figure}[t]
	\centering
	\includegraphics[width=.3\textwidth]{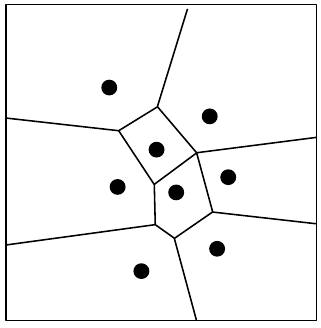}
	\caption{An example of a Voronoi diagram using 8 codeword points.}
	\label{fig:voronoi}
\end{figure}

Nearest neighbor quantization, or Voronoi quantization, is an encoding technique to group sets of similar points from a dataset $X$ \cite{gray_1984_VectorQuantization}. Each region is defined by a codeword point $c \in \Re^n$ that forms a region where each other point $x \in \Re^n$ in that region is closest to $c$ than any other codeword. These regions take on the form of $m$ disjoint Voronoi cells following 

\begin{align}
	R_i = \left\{ x \in \Re^n: \psi \left ( x \right ) = c_i \right\}
\end{align}

\noindent
for any $i = 1, 2, .., m$ and quantizer $\psi$. The mapping of a point $x$ onto a region is defined by

\begin{align}
	R_i = \left\{ x: \left\| x - c_i\right\| \leq \left\| x - c_j\right\|\right\}, \forall j \neq  i
\end{align}

\noindent
with $c_i$ as the nearest neighbor codeword of $x$. The resulting regions look like cell-like structures, that rely on a distance measure for classification, instead of a parameterized decision boundary. This compresses the original dataset onto a finite set of codewords $C = \left\{c_1, c_2, ..., c_m \right\} \subset \Re^n$.

A main benefit of Voronoi quantization is its simple representation, with codeword $C$ being a subset of the dataset. The distance-based metric to find a nearest neighbor is computational cheap and is easily understandable when inference is done relative to all other neighboring regions.
The mapping from point to region is performed using a kd-tree \cite{bentley_1975_MultidimensionalBinary}, which is an efficient partitioning datastructure for k-dimensional spaces. In average case, when the tree is balanced, the lookup of a region exhibits logarithmic time complexity.

\paragraph{\textbf{k-means clustering}}

K-means clustering is an unsupervised learning approach where a dataset is partitioned into $k$ clusters \cite{macqueen_1967_MethodsClassification}. It extends on Voronoi quantization by specifying the number of clusters in the model upfront. The assignment of any point to a cluster is done by looking up the centroid or mean of the cluster that has the lowest distance to that point. 

Learning the partitioning of a dataset $\left( x_0, x_1, ..., x_n \right)$, and therefore the $k$ centroids, is iteratively done by minimizing the within-cluster sum of squares (WCSS) given by

\begin{align}
	\textrm{argmin}_R \sum_{i=0}^{k}\sum_{x\in R_i}\left\| x - c_i \right\|
\end{align}

\noindent
where each centroid $c_i$ is the mean of the cluster following

\begin{align}
	c_i = \frac{1}{\left| R_i \right|}\sum_{x \in R_i} x
\end{align}

If $k$ is not known upfront, an optimal value can be found by going over a range of k-values, create a corresponding model for each candidate and calculate the silhouette score of each model. This is a measure of cohesion (how close points are within their cluster) and separation (how far is each point from the closest neighbor cluster) combined into a value between -1 and 1. A positive score indicates a good clustering of the points while negative values indicate insufficient placement of the centroids.
Given dataset $x$ that part of cluster $R_i$, we define the silhouette score as

\begin{align}
	s(x) = \frac{b(x) - a(x)}{\max\left\{ a(x), b(x) \right\}}
\end{align}

\noindent
where $a(x)$ is the average distance from $x$ to all points in cluster $R_i$, $b(x)$ is the average distance score of all points in the nearest cluster and $|R_i| > 1$. The highest scoring $k$ is the one that with an optimal amount of clusters.
We will use this method in out algorithm to find an ideal amount of new regions per codeword and how their codewords would be defined.

\section{Related work}

With background on Reinforcement Learning, Voronoi quantization and k-means laying the foundation to our method, we now review existing approaches at distilling a Deep RL policy in a more interpretable model.

\subsection{Distillation into interpretable models}
A well known technique to generate a local explanation for a classification model is Local Interpretable Model-agnostic Explanations (LIME) \cite{ribeiro_etal_2016_WhyShould}. It involves fitting a linear function at a point in the input space that will stand in for an explanation how the model comes to a decision. Around a point of interest, a set of nearby samples is taken with the distance of each sample to the point counted as a weighing factor. The closer a sample is, the more it should contribute to the explanation since it is more similar to the point of interest. On this weighted set, a linear function is fitted using linear regression. Afterwards, the weights of the linear function indicate the importance of each input variable to decide the output of the model \cite{barredoarrieta_etal_2020_ExplainableArtificial}. 
As the acronym implies, it is also applicable for policy networks that perform a classification or regression task within the action space. For example, LIME has been used by Gjaerum et al.  \cite{gjaerum_etal_2021_ExplainingDeep} to explain the docking policy of an autonomous maritime vehicle.

For global policy descriptions, a hierarchical model such as a decision tree reduces the complexity of an explanation since it takes a single path of decisions to come to the action of a single situation. The traditional decision rules nodes and fixed output leaves can be replaced by linear models to increase their capacity. Liu et al. \cite{liu_etal_2019_InterpretableDeep} explain behaviors of simple control problems using this approach, names in their case Linear Model U-Tree (LMUT). Green \cite{green_sheppard_2024_PerformanceRobustness} further refine the approach by introducing pruning.
Kohler et al. \cite{kohler_etal_2024_InterpretableEditable} use oblique trees with their INTERPRETER method to distill policies for feature-extracted Atari games. An oblique tree performs splits oblique to the axis of the input variables, reducing the depth of the tree. However, the oblique cuts are more difficult to read by a human (compared to simple $x > y$ conditions).
Coppens et al. \cite{coppens_etal_2019_DistillingDeep} applied knowledge distillation on soft decision trees (SDT) to learn a policy in a tile-based game. Here, the tree is a fixed structure where the decision nodes contain a single perceptron. Each leaf contains a softmax distribution over the discrete actions. A decision made by an SDT incorporated all possible paths to the leaf nodes rather than one greedy path towards an action. 

\subsection{Space partitioning}

Another post-hoc approach is to simplify the complexity in the state space of the policy. By clustering group of states together, according to a similarity measure, a partitioning of local explanations can give insights on abstract situations.

One early example, and main inspiration for out approach, is TD-AVQ by \cite{lee_lau_2004_AdaptiveStatea}. Here, adaptive state partitioning, using Voronoi quantization, is proposed as a solution to enable tabular Temporal Difference learning to be done in continuous state space. Their main motivation was not Explainable RL. It was, at the time (before efficient neural networks), a necessity for efficiently handling continuous states in Reinforcement Learning.

Akrour et al. cluster the state space in regions where subpolicies of a simple shape produce near-optimal behaviour \cite{akrour_etal_2018_RegularizingReinforcement}. Their method decomposes the original policy and state space using a neural networks that acts as the upper level policy, responsible for assigning states to a set of linear policies. The use of linear policies produces acceptable explanations, but the top-level neural-network policy is not explainable, leading to only modest explainability of their full approach.

The approach we propose in this paper combines the explainable Voronoi partitioning of Lee et al. with the local linear policies of Arkour et al., and with a special distillation algorithm that considers not only the difference between black-box and distilled actions, but also the impact of these differences on performance (using learned Q-Values).

\section{Partitioning the state space}
In this section, we discuss how we approach the partitioning of a black box policy into a set of simpler models using a Voronoi quantizer. We accomplish this by alternating between optimizing the current set of subpolicies and adding new ones for states where the associated subpolicy cannot produce behavior that is sufficiently close to the original policy. We motivate the use of the critic value network from the original DRL policy and how it guides the selection process of new candidate codewords.

We give an overview of the algorithm in Alg. \ref{alg:vsp}. It runs for $\texttt{n\_iteration}$ iterations and requires both a set of state-action pairs and the critic from the original policy.

\begin{algorithm}
	\caption{Critic-driven Voronoi State Partitioning}
	\begin{algorithmic}[1]
		
		\Require State-action pairs $\left\langle S_\pi, A_\pi \right\rangle$, critic network $Q_\pi$, environment \texttt{env}
		\State Initialize list of subpolicies $\tilde{\pi}$ with arbitrary policy $\tilde{\pi}_0$ and $s_0$ as codeword $c_0 \in C$
		
		\For{$n = 0$ to $\texttt{n\_iteration}$} \Comment{Early stopping on performance possible}
		
		\State Distribute $\left\langle S_\pi, A_\pi \right\rangle$ over $\tilde{\pi}$ using $\psi_C$ 
		
		\Repeat
		\State Train each subpolicy $\tilde{\pi}_i$ on batches of $\left\langle S_\pi, A_\pi \right\rangle_i$
		\Until Early stopping
		
		\ForAll{$\tilde{\pi}_i$}
        \State $\dot{S}_{\tilde{\pi}_i} \gets$ every $s \in S_{\tilde{\pi}_i}$ s.t. ${\left\| s, c_{\tilde{\pi}_i} \right\|}_2 > D_{-}$
		\State $X \gets$ the top-$N_t$ \% elements of $\dot{S}_{\tilde{\pi}_i}$ by $Q_\pi\left(s, \tilde{\pi_i}\left( s\right)  \right)$
        
		\ForEach{$\dot{c} \in \operatorname{findClusters}\left( X \right)$}
		\State Add $\dot{c}$ to $C$
        \State Break to $\tilde{\pi}_{i+1}$ if we added more than $N_r$ (per-region) codewords
        \State Break to $n+1$ if we added more than $N_i$ (per-iteration) codewords
		\EndFor
		
		\EndFor
		\EndFor

	\end{algorithmic}
	\label{alg:vsp}
\end{algorithm}

\begin{algorithm}
	\caption{findClusters}
	\begin{algorithmic}[1]
		
		\Require Dataset of states $X$
		\State Initialize list of scores $S$ and centroids $C$
		\For{$m=2$ to $\texttt{max\_k\_clsuters}$}
		\State Create m-means model $F_m$ using $X$
		\State Calculate silhouette score $S_{F_m}\left( X \right)$
		\State Add $S_{F_m}$ to $S$
		\State Retrieve centroids from $F_m$ and add to $C$
		\EndFor
		\State $n = \operatorname{argmax}_m\left( S \right) $
		\State Return centroids $C_n$

	\end{algorithmic}
	\label{alg:findClusters}
\end{algorithm}

\subsection{Training subpolicies}
The algorithm requires a dataset of experiences from an original policy $\pi$ to be used as the training set for the set of subpolicies $\tilde{\pi}$. This consists of state-action pairs $\left\langle s_\pi, a_\pi \right\rangle$ that were collected in the rollout of $\texttt{n\_episodes\_training}$ episodes with the original policy in the environment. This dataset captures the mapping of the black box we want to mimic.

A subpolicy is a parameterized model that maps a state $s$ to an action $a$. In this paper, we limit ourselves to linear functions where each variable $a_k$ of an action is given by

\begin{align}
	a_k = \sum_{i=0}^{n}\left( s_i * w_{i,k} \right) + b_k
\end{align}

\noindent
with weights $w$ and biases $b$. Any other class of parameterized model is allowed, but we focus in this paper on this model, favoring explainability over expressiveness.

We define a region $r_i \in R$ defined by Voronoi quantizer $\psi_C$ as a pair $\left\langle c_i \tilde{\pi}_i \right\rangle$  that associates a codeword $c_i$ to a subpolicy $\tilde{\pi_i}$. At initialization, the partitioning consists of only one region $\left\langle c_0, \tilde{\pi}_0 \right\rangle$. The first codeword $c_0 = s_0$ is equal to the first state $s_0$ of the first episode in the dataset. The parameters of the initial subpolicy $\tilde{\pi}_0$ are set arbitrarily.

In the first iteration of the algorithm the entire dataset is used to optimize $\tilde{\pi}_0$. Optimizing a subpolicy is done via gradient descent (in our case using the Adam optimizer) using mini-bathes for a maximum of $\texttt{n\_epochs}$. After each epoch, an early stopping condition checks if the training loss of the subpolicy decreased with a minimum delta threshold over a patience window of epochs. Because we are fitting a linear model, we do not foresee overfitting to be an issue.

After optimization, the mean of the training loss is stored and used later to decide which regions should be split first. The intuition behind this is that the training loss indicates how far a policy could be optimized.

\subsection{Defining new regions}
New regions, and therefore new codewords, should be introduced within subsets of states of a region $R_i$ where the current subpolicy $\tilde{pi}_i$ is unable to produce behavior close to the original policy $\pi$. To accomplish this, we opt for a selection procedure following eligibility criteria that define which states could be part of a region and which ones should be split off. 

The first criterium for a state to be excluded from a region $R_i$ is when it is situated far enough in state space from the centroid of that region. Instead of deciding upon a fixed number of subpolicies upfront, our approach is set to a certain minimum resolution equal to $\texttt{min\_codeword\_distance}$ which it the minimum distance between any state of one region to the centroid codeword of another. For each state $s$ contained in region, we check wether or not the threshold is met when calculating Euclidean distance ${\left\| s, c_{\tilde{\pi}_i} \right\|}_2$. States with a distance larger than this threshold are accumulated in the set of candidate states $\dot{S}$.\\

The second criterium for a state belonging to a new region is based on the value predicted by the critic $Q_{\pi}$ of the original policy. For each candidate $s \in \dot{S}$, we compute the action $a = \tilde{\pi_i}\left( s \right)$ and evaluate this state-action pair using the critic. The result is $q_s = Q_\pi\left(s, \tilde{\pi_i}\left( s\right)  \right)$ which is the value of the predicted action of the subpolicy in that state.
With all candidate states evaluated, we sort $\dot{S}$ in increasing order of value and take a proportion $X = \dot{S}_{[0, \rho]}$ of states with $\rho = \lfloor \texttt{value\_ratio\_threshold} * |\dot{S}| \rfloor$. These states will be considered of low enough value, and therefore bad enough in what actions the subpolicy predicts. This is an analogue for states tat require behaviour too complex for the class of subpolicy we use.\\

Given the set $X$, we use the function $\operatorname{findClusters}$ to find a suitable set of codewords $\dot{C}$ that define new regions based on state similarities (Alg. \ref{alg:findClusters}). This function will construct different k-mean models with $k \in [2, .., \texttt{max\_k\_clusters}]$ and calculate the silhouette score for each model. The model with the highest score is considered as the best clustering and the centroids of this model are assigned to $\dot{C}$. For each centroid, we append it as a codeword to $C$ until the maximum amount of codewords per region $\texttt{max\_codewords\_region}$ is met. This parameters controls how fast a region is being subdivided in each iteration. For each codeword we add, we extend the list of subpolicies $\tilde{\pi}$ with a subpolicy with arbitrarily set parameters. \\

We execute the procedure above for each codeword in $C$ or until the maximum amount of newly introduced codewords $\texttt{max\_codewords\_iteration}$ is achieved.We execute the algorithm for $\texttt{n\_iterations}$ or when the amount of found regions converges. 
Since the new subpolicies are trained in the iteration after they are introduced, we have to ensure that the last iteration of the algorithm is one that does not extend the regions. This can simply be done by looking at when the algorithm converges and take one iteration after it as the final result.







\section{Experimental validation}

In this section, we validate our partitioning algorithm on both the performance on each control task as well as the kind of insights we get from the learned subpolicies.

We start by describing the used environments, which are all well-known continuous control benchmarks from the Gymnasium module except one. Secondly we examine the results of running the algorithm for each of these environments and evaluate the performance in terms of mean episodic return. For all environments, we seek to find sets of linear functions that imitate the behavior of a trained DRL policy.
Finally, we look at several partitionings, what subpolicies have been found and what what insight we can retrieve from them.\\

For each of the environments a TD3 policy was trained using optimal hyperparameters from StableBaselines3 Zoo. For each environment, we ran 20 runs of both our critic driven Voronoi State Partitioning (VSP-critic) as well as one variant of VSP that chooses new codewords randomly (VSP-random). For each run we generate a new dataset containing 100 episodes of state-action pairs with our TD3 policy and we extract the original critic from it. The evaluation of the RL policy is an average over all episodes that constitute each dataset.

The used hyperparameters are listed in Table \ref{tab:hyperparems}. These are hand-tuned to have a good balance between performance and the amount of subpolicies found and are not guaranteed te be optimal. 

\begin{table}[]
	\begin{tabular}{|c|c|c|c|c|c|}
		\hline
		Environment   & $\texttt{min\_c\_dist}$ & $\texttt{val\_ratio}$ & $\texttt{max\_c\_region}$ & $\texttt{max\_c\_iteration}$ & $\texttt{n\_iterations}$ \\ \hline
		SimpleGoal    & 0.6                     & 0.5         			& 2                         & 3                            & 10             		  \\ \hline
		MountainCar   & 0.1                     & 0.8       		    & 3                         & 5                            &  20                      \\ \hline
		LunarLander   & 0.35                    & 0.6        			& 3                         & 5                            &  50                      \\ \hline
		BipedalWalker & 1                       & 0.5       		    & 1                         & 10                           &  100                     \\ \hline
	\end{tabular}
	\caption{Used hyperparameters for comparing both VSP-critic and VSP-random. C stands for codeword abbreviated.}
	\label{tab:hyperparems}
\end{table}

\subsection{Validation environments}

\begin{figure}[t]
	\centering
	
	\begin{subfigure}[b]{0.24\textwidth}
		\centering
		\adjustbox{valign=b}{\includegraphics[height=1.8cm]{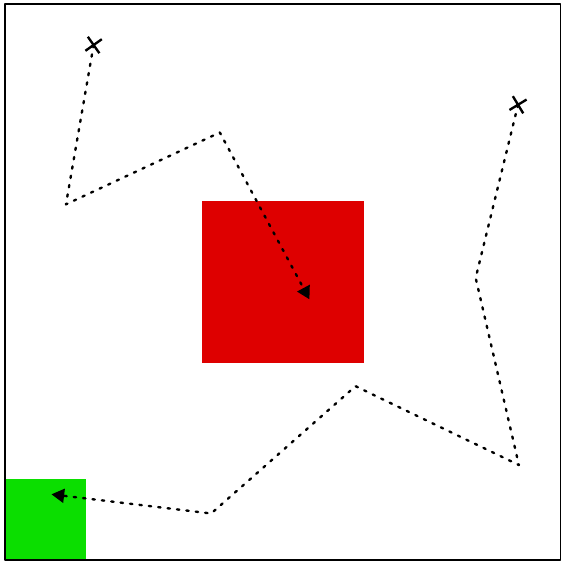}}
	\end{subfigure}
	\hfill
	\begin{subfigure}[b]{0.24\textwidth}
		\centering
		\adjustbox{valign=b}{\includegraphics[height=1.8cm]{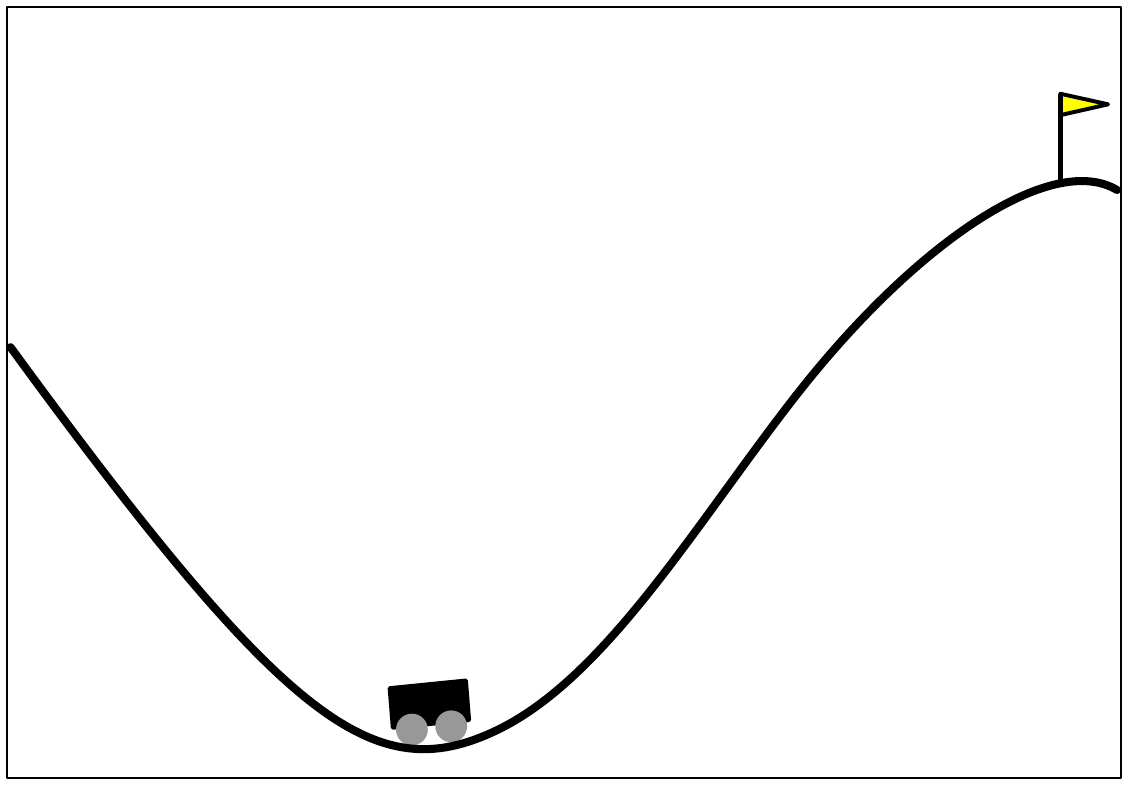}}
	\end{subfigure}
	\hfill
	\begin{subfigure}[b]{0.24\textwidth}
		\centering
		\adjustbox{valign=b}{\includegraphics[height=1.8cm]{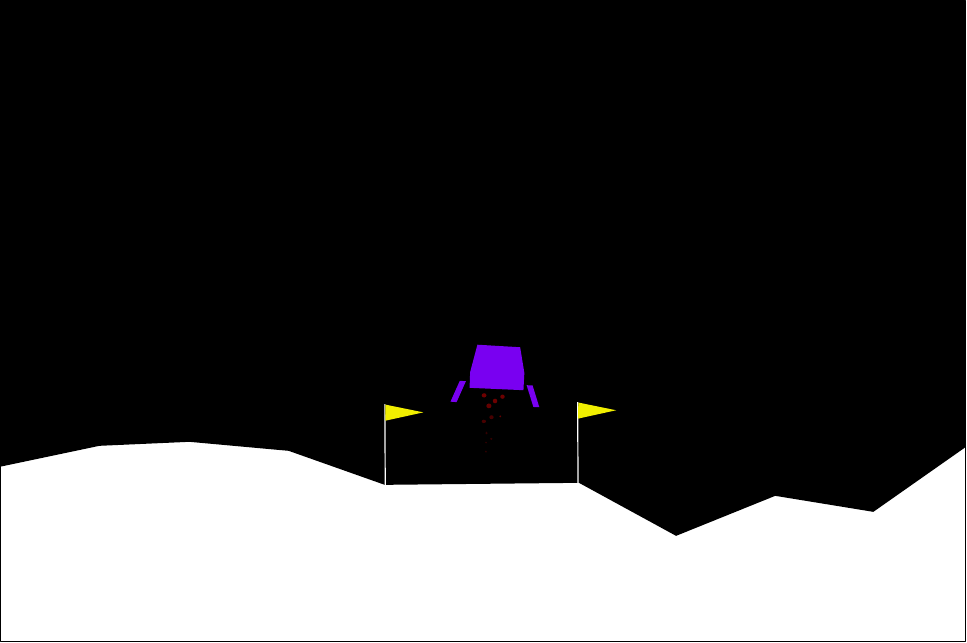}}
	\end{subfigure}
	\hfill
	\begin{subfigure}[b]{0.24\textwidth}
		\centering
		\adjustbox{valign=b}{\includegraphics[height=1.8cm]{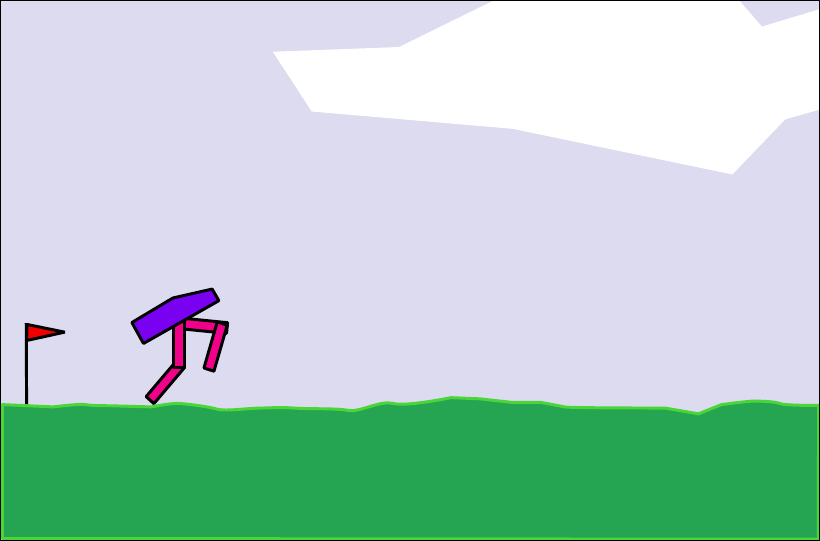}}
	\end{subfigure}
	
	\caption{The used validation environments. From left to right: SimpleGoal, MountainCarContinuous, LunarLanderContinuous and BipedalWalker.}
	\label{fig:environments}
\end{figure}

\paragraph{\textbf{SimpleGoal}}In this navigation task, the agent has to reach the goal state (green) from a random initial state while avoiding a pitfall region (red) as fast as possible. The environment is a bounded continuous space $1.0 \times 1.0$ with the goal state at $x < 0.1, y < 0.1$ and the pitfall at $0.4 < x < 0.6, 0.4 < y < 0.6$. The observation space consists of the current $x$ and $y$ coordinate of the agent and the action is the change in $x$ and $y$ for the next step. The applied step in the environment is scaled by a factor $0.1$. A potential based reward $r_t = 10 * (\texttt{old\_distance} - \texttt{new\_distance})$ is given at each timestep which indicate the progress the agent makes towards the goal. Stepping in the pitfall terminates the episodes and returns a reward of $-10$ while reaching the goals earns the agent a reward of 10. An episode is truncated at 50 steps.

The three other environments, \textbf{MountainCarContinuous}, \textbf{LunarLanderContinuous} and \textit{BipedalWalker}, are described by \cite{towers_etal_2024_GymnasiumStandard}.

\subsection{Distillation performance}
\begin{figure}[t]
\begin{minipage}{0.84\textwidth}
	\begin{minipage}[h]{0.5\linewidth}
		\centering
		\includegraphics[width=1\linewidth]{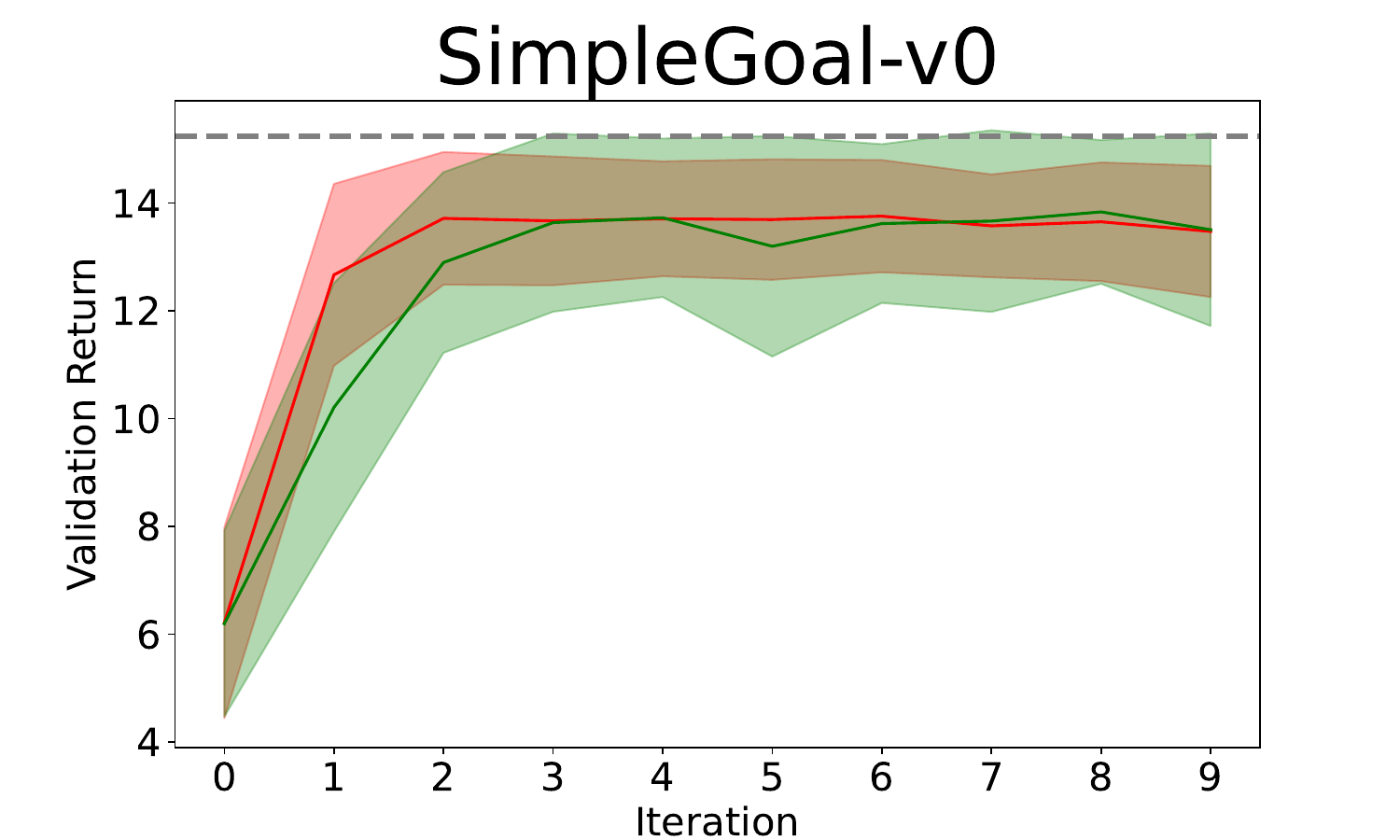}
	\end{minipage}
	\hfill
	\begin{minipage}[h]{0.5\linewidth}
		\centering
		\includegraphics[width=1\linewidth]{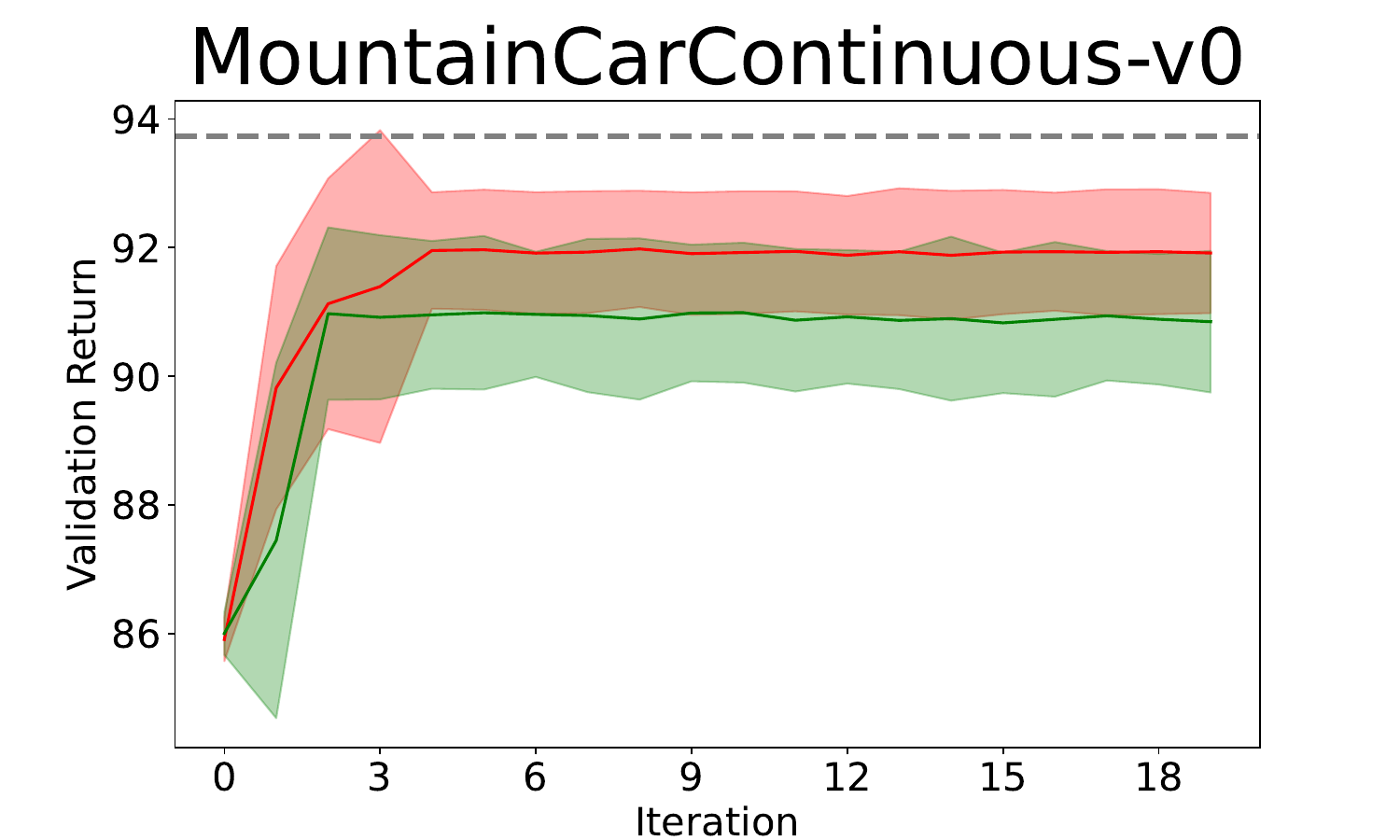}
	\end{minipage}
	\vfill
	\begin{minipage}[h]{0.5\linewidth}
		\centering
		\includegraphics[width=1\linewidth]{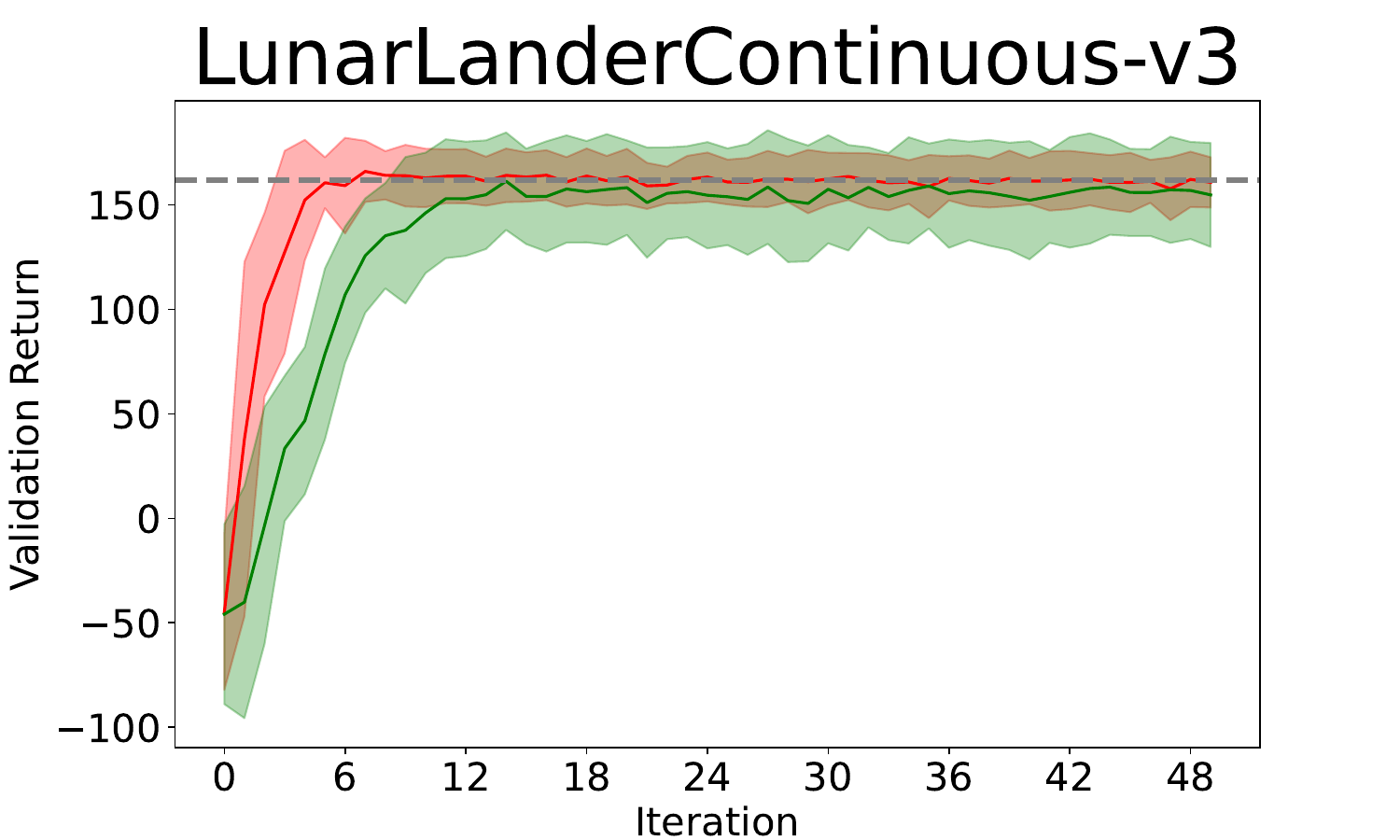}
	\end{minipage}
	\hfill
	\begin{minipage}[h]{0.5\linewidth}
		\centering
		\includegraphics[width=1\linewidth]{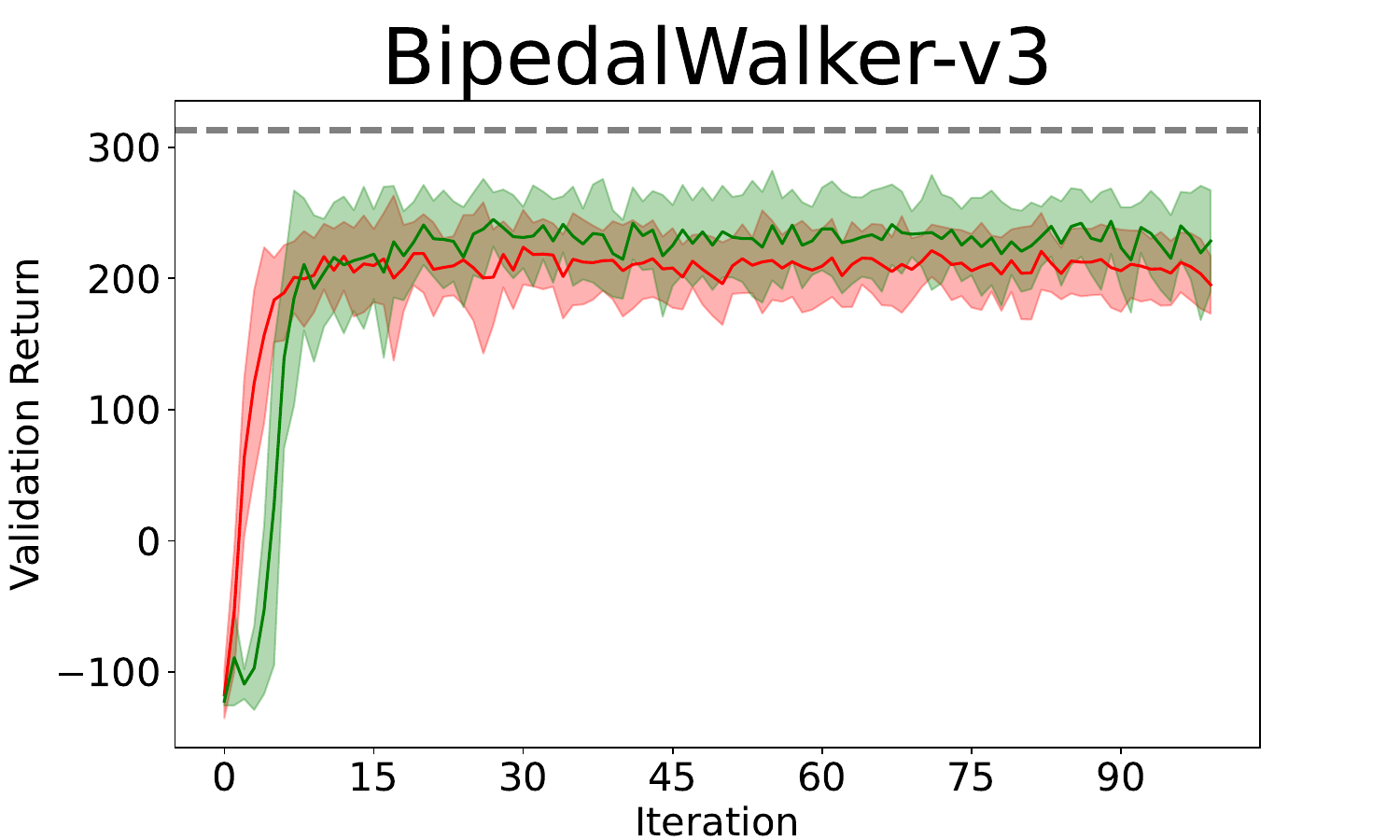}
	\end{minipage}

\end{minipage}
\begin{minipage}{0.15\textwidth}
	\centering
	\includegraphics[width=1\linewidth]{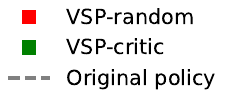}
\end{minipage}
\caption{Performance of both the VSP-critic algorithm (green) and the random variant (red) in function of iterations of the algorithm. Each plot represents the evolution of the mean collected return per environment over 20 executions of the methods. The grey dotted line is the average episodic return of the original TD3 policy. The area around the curves is 1 standard deviation from the mean.}
\label{fig:iteration_performances}
\end{figure}

The average mean return of both VSP-critic and VSP-random are given in Table \ref{tab:performances}. We observe that for all environments except BipedalWalker the average return of each algorithm approaches the original policy. VSP-random outperforms VSP-critic with a small difference in all environments but BipedalWalker where the critic-driven version clearly outperforms random. Often the spread in performance among VSP runs is higher compared to TD3, which is especially noticeable for VSP-critic. The performance of the subpolicies learned over the iterations of the algorithm, shown in Figure \ref{fig:iteration_performances} indicate that both algorithms converge relatively quick. We observe that changing the hyperparameters that govern the introduction of new regions ($\texttt{max\_codewords\_region}$ and $\texttt{max\_codewords\_iteration}$) have an impact on when this convergence happens.

\begin{figure}[]
\begin{minipage}{0.84\textwidth}
	\begin{minipage}[h]{0.5\linewidth}
		\centering
		\includegraphics[width=1\linewidth]{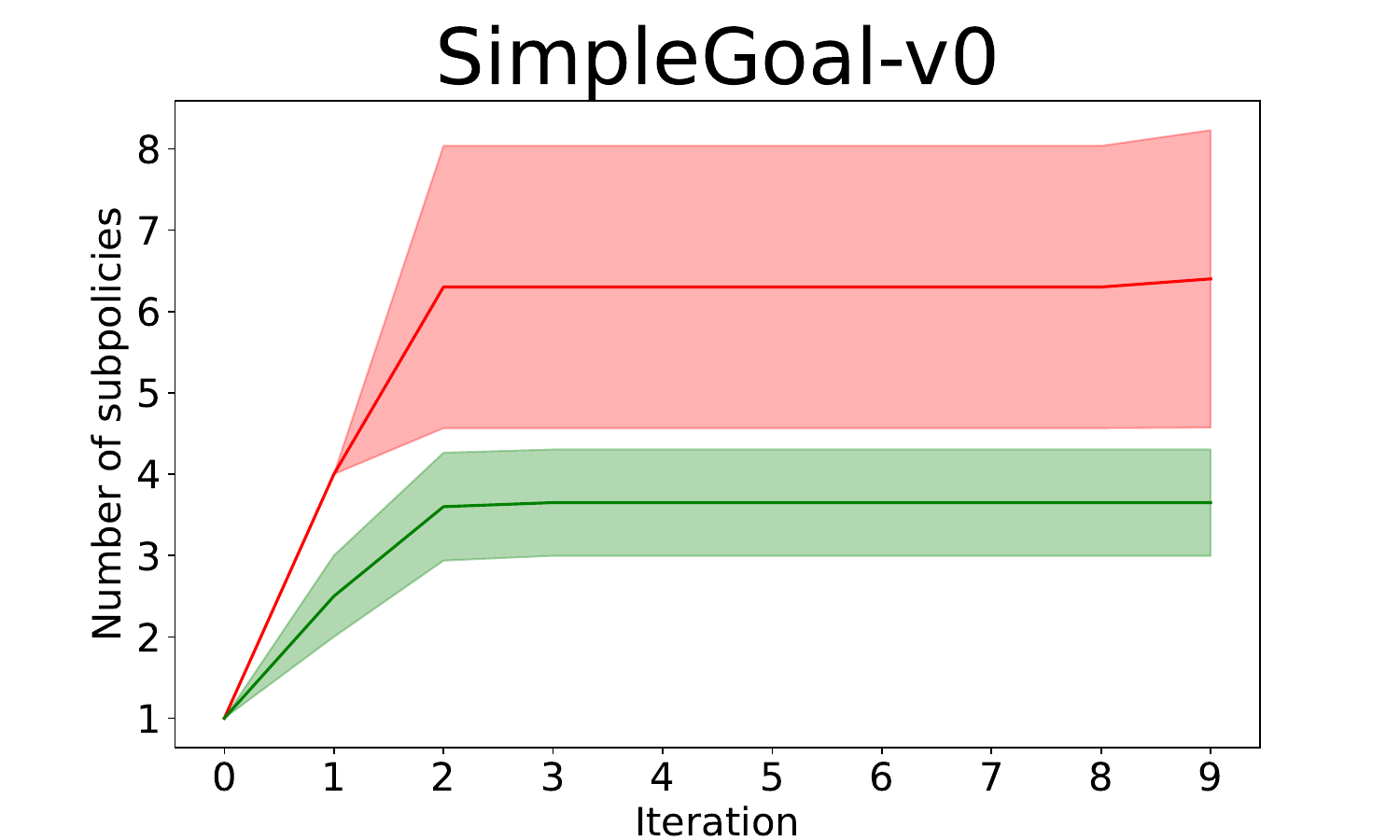}
	\end{minipage}
	\hfill
	\begin{minipage}[h]{0.5\linewidth}
		\centering
		\includegraphics[width=1\linewidth]{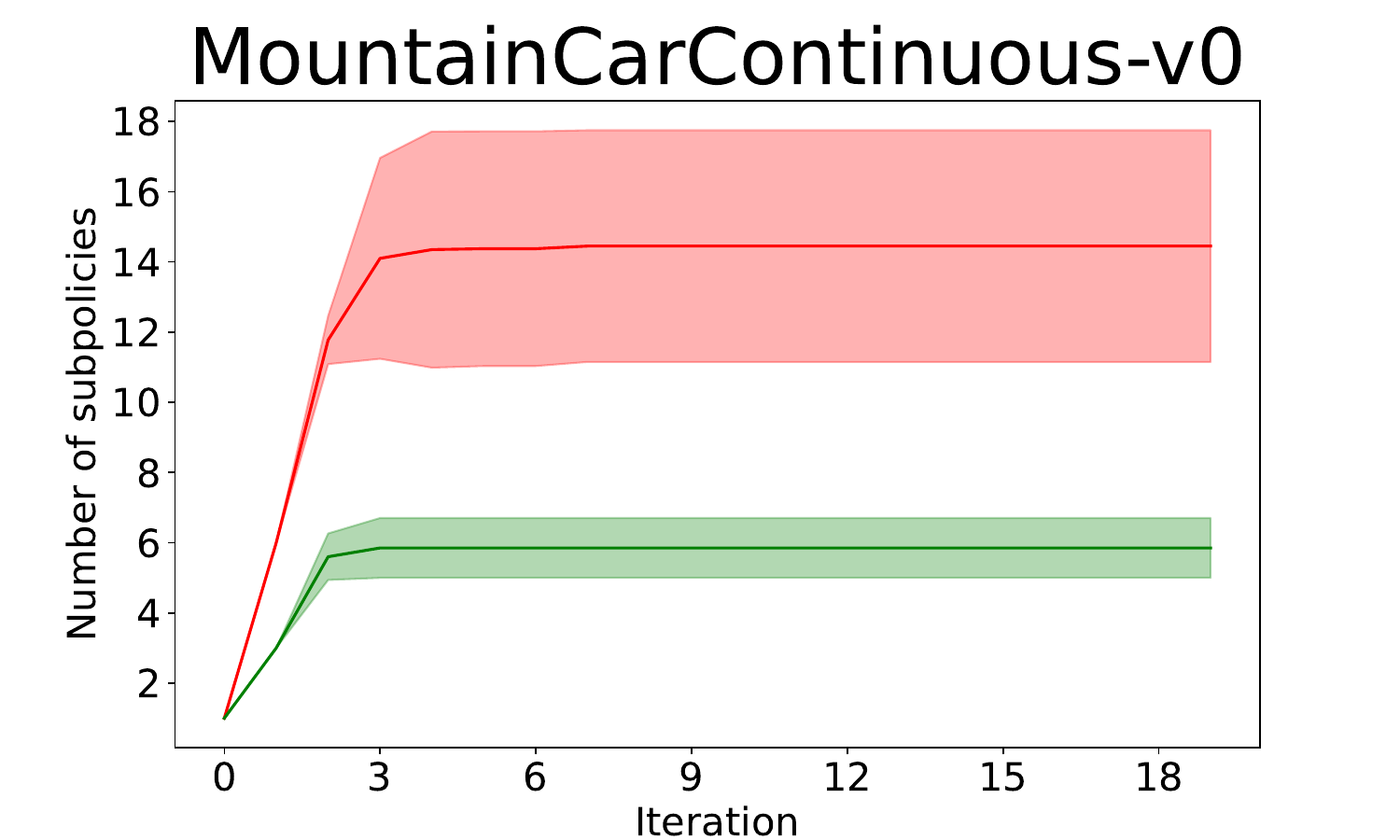}
	\end{minipage}
	\vfill
	\begin{minipage}[h]{0.5\linewidth}
		\centering
		\includegraphics[width=1\linewidth]{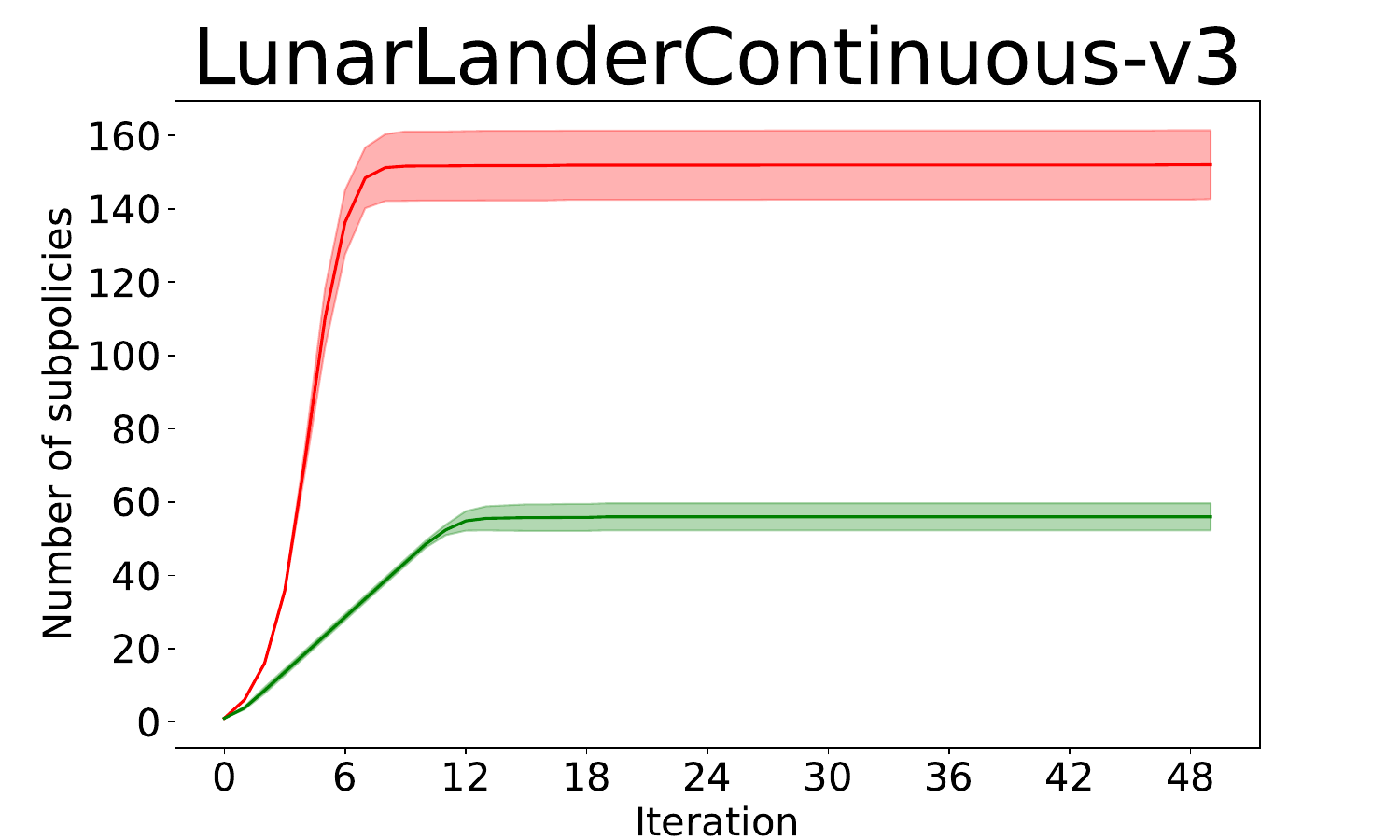}
	\end{minipage}
	\hfill
	\begin{minipage}[h]{0.5\linewidth}
		\centering
		\includegraphics[width=1\linewidth]{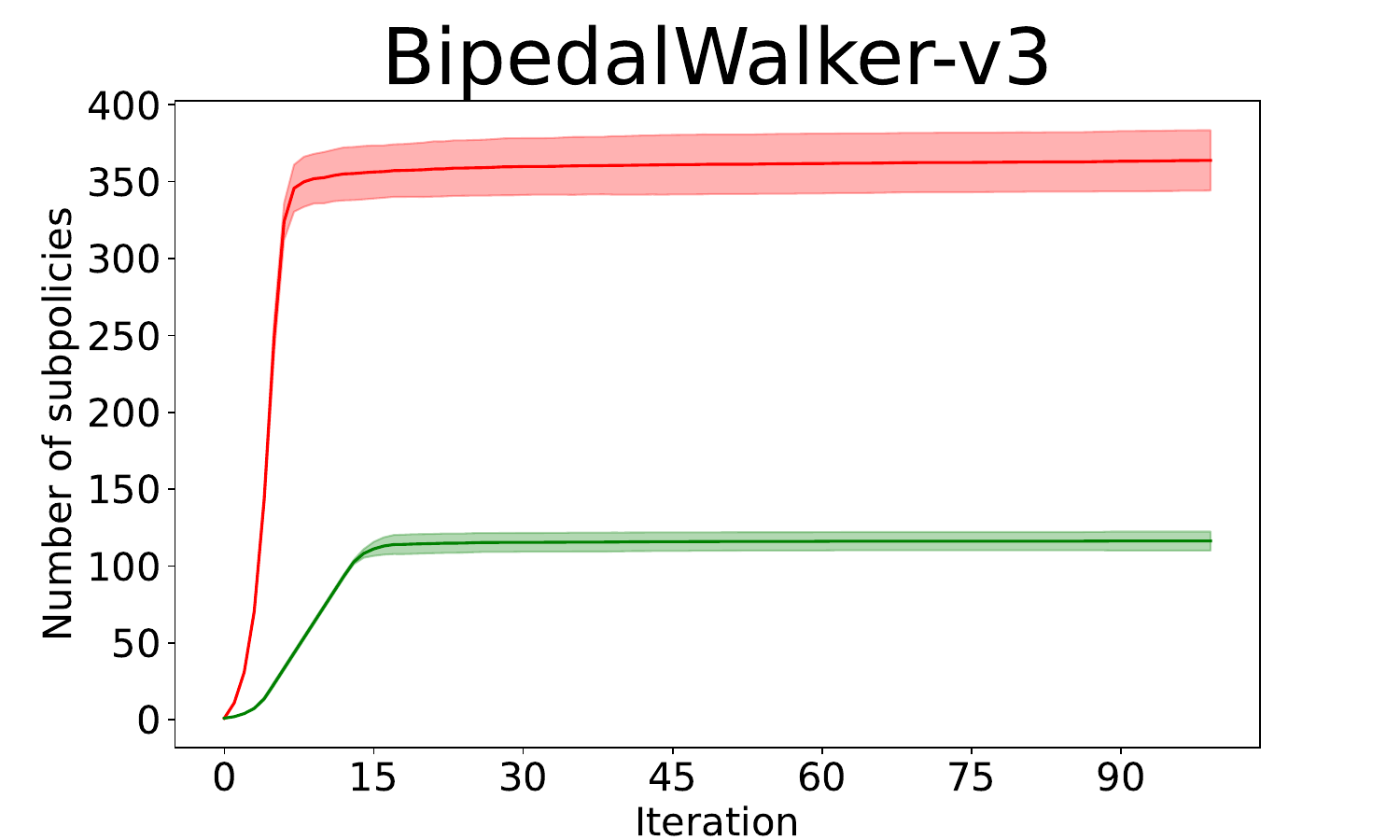}
	\end{minipage}
	
\end{minipage}
\begin{minipage}{0.15\textwidth}
	\centering
	\includegraphics[width=1\linewidth]{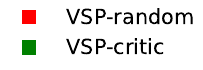}
\end{minipage}
\caption{Number of linear subpolicies found by the VSP algorithm (green) and the random version (red). The area around the curves is 1 standard deviation from the mean.}
\label{fig:n_policies}
\end{figure}

If we look at the number of subpolicies learned, we see that VSP-critic only needs a fraction of the subpolices VSP-random needs to achieve similar results. From Figure \ref{fig:n_policies} we can see that here that the amount of subpolicies convergences early as well. For example, VSP-critic needs around 60 linear functions to control LunarLander decently while VSP-random needs around three times that amount. In the case of BipedalWalker, it clearly indicates that having more subpolicies does not translate into higher performance.

\begin{table}[]
	\centering
	\begin{tabular}{|c|c|c|c|c|c|}
		\hline
		Environment   & $\operatorname{Return}_{\operatorname{TD3}}$      & $\operatorname{Return}_{\operatorname{critic}}$ & $\operatorname{Return}_{\operatorname{rand}}$ & $\left| \tilde{\pi}_{\operatorname{critic}} \right| $ & $\left| \tilde{\pi}_{\operatorname{rand}} \right| $ \\
		\hline
		SimpleGoal    & $15.2 \pm 0.13$ & $13.5 \pm 1.79 $  & $ 13.4 \pm 1.22$  & $3.6 \pm 0.65$  & $6.4 \pm 1.83$      \\
		\hline
		MountainCar   & $93.7 \pm 0.02$ &  $90.8 \pm 1.10$  & $ 91.9 \pm 0.93$  & $5.8 \pm 0.85$  & $14.4 \pm 3.30$     \\
		\hline
		LunarLander   & $161.8 \pm 8.28$&  $154.8 \pm 24.91$& $160.8 \pm 12.04$ & $55.9 \pm 3.71$ & $152.0 \pm 9.41$    \\
		\hline
		BipedalWalker & $313.0 \pm 4.14$& $226.0 \pm 39.41$  & $195.1 \pm 21.96$& $116.7 \pm 6.36$& $363.8 \pm 19.64$   \\
		\hline     
	\end{tabular}
	\caption{Average return over 20 runs for TD3, VSP-critic and VSP-random as well as the number of the found subpolicies.} 
	\label{tab:performances}
\end{table}

\subsection{Voronoi diagram}
\begin{figure}[t]
\centering
	\includegraphics[width=0.32\textwidth]{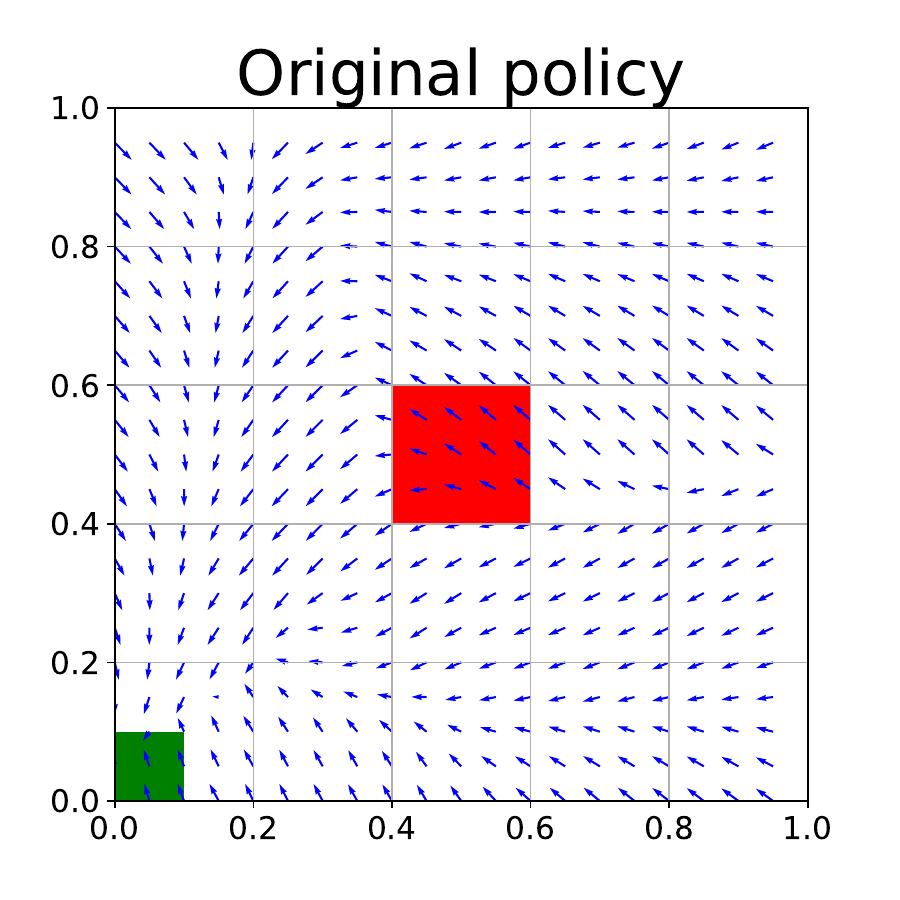}
	\hfill
	\includegraphics[width=0.32\textwidth]{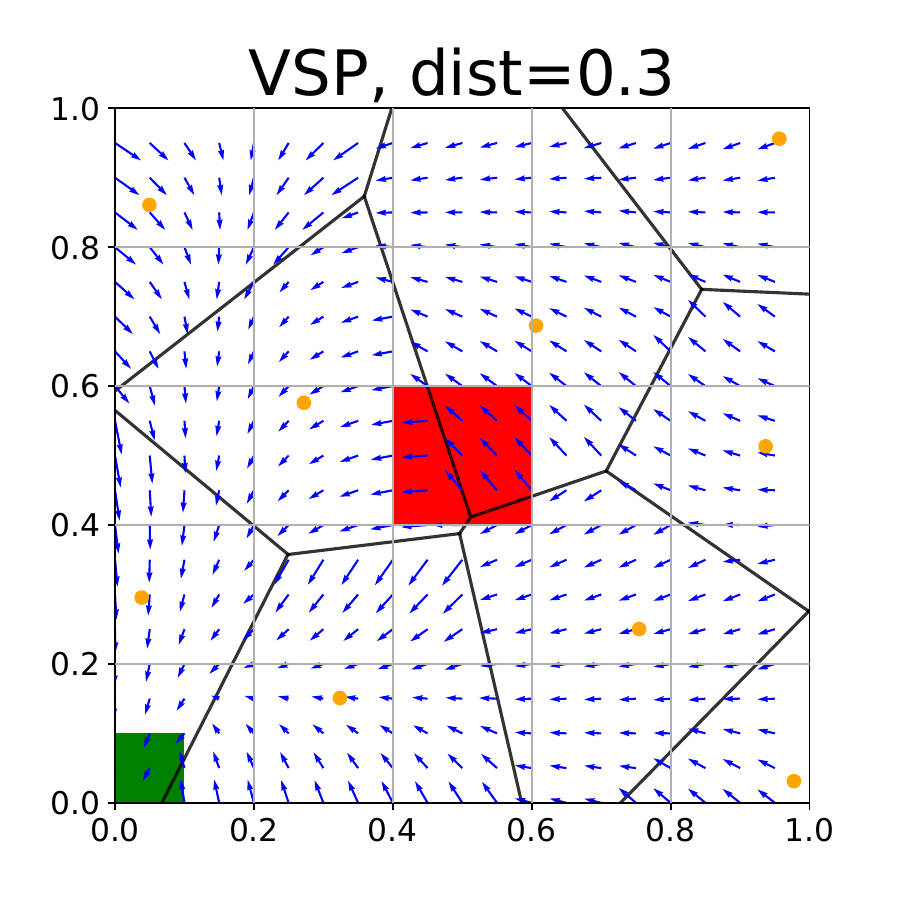}
	\hfill
	\includegraphics[width=0.32\textwidth]{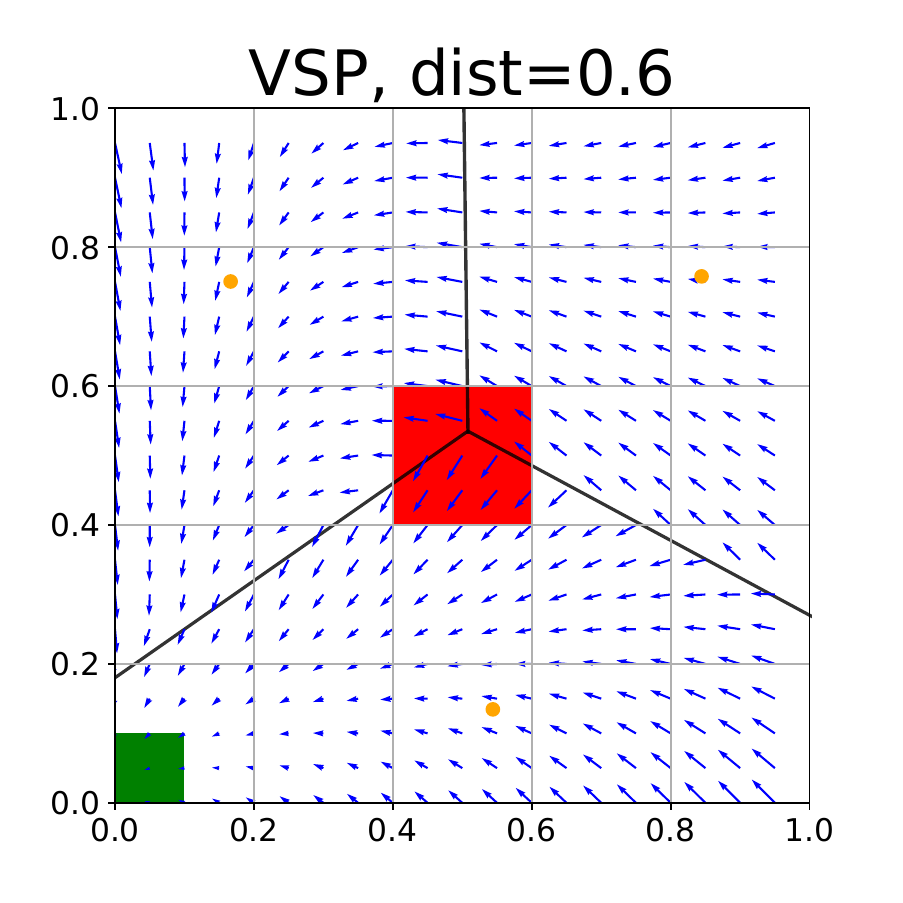}

	\caption{Control policies for the SimpleGoal environment using quiver plots. Each arrow indicates the direction and step size the policy would take in that state. When comparing two sets of subpolicies (middle and right) with different $\texttt{min\_codeword\_distance}$, wee see that the behaviours are similar to the original TD3 policy (on the right). We count 9 regions for a distance of $0.3$, while only 3 are found if the minimum distance is set to $0.6$}
	\label{fig:simplegoal_quiver}
\end{figure}

\begin{figure}[]
	\centering
	
	\begin{subfigure}[t]{0.48\textwidth}
		\centering
		\includegraphics[width=\linewidth]{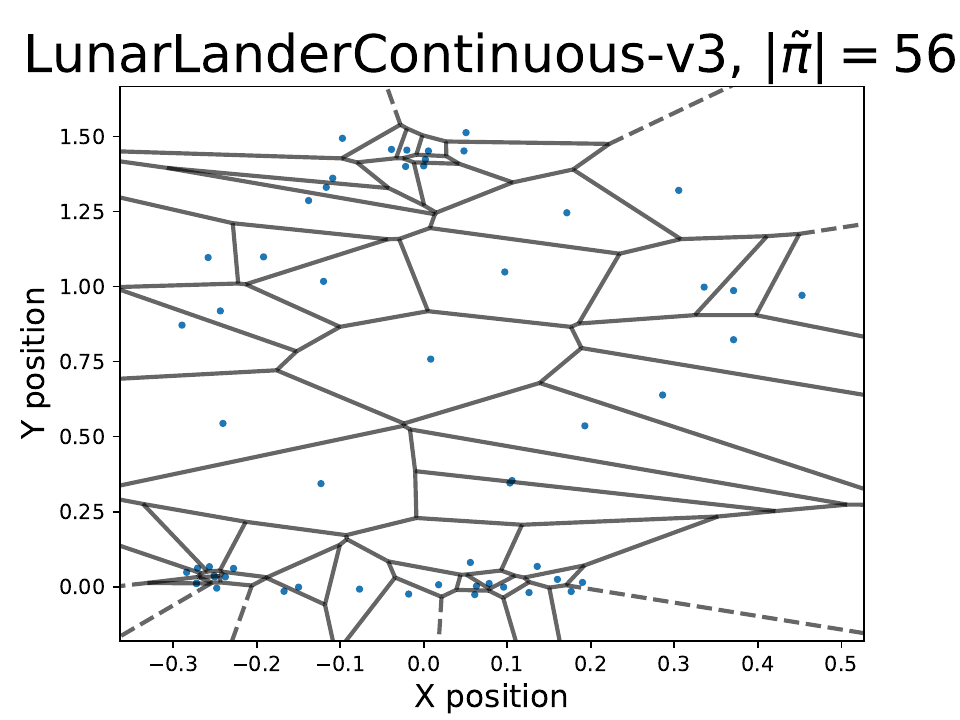}
		\label{fig:first}
	\end{subfigure}
	\hfill
	\begin{subfigure}[t]{0.48\textwidth}
		\centering
		\includegraphics[width=\linewidth]{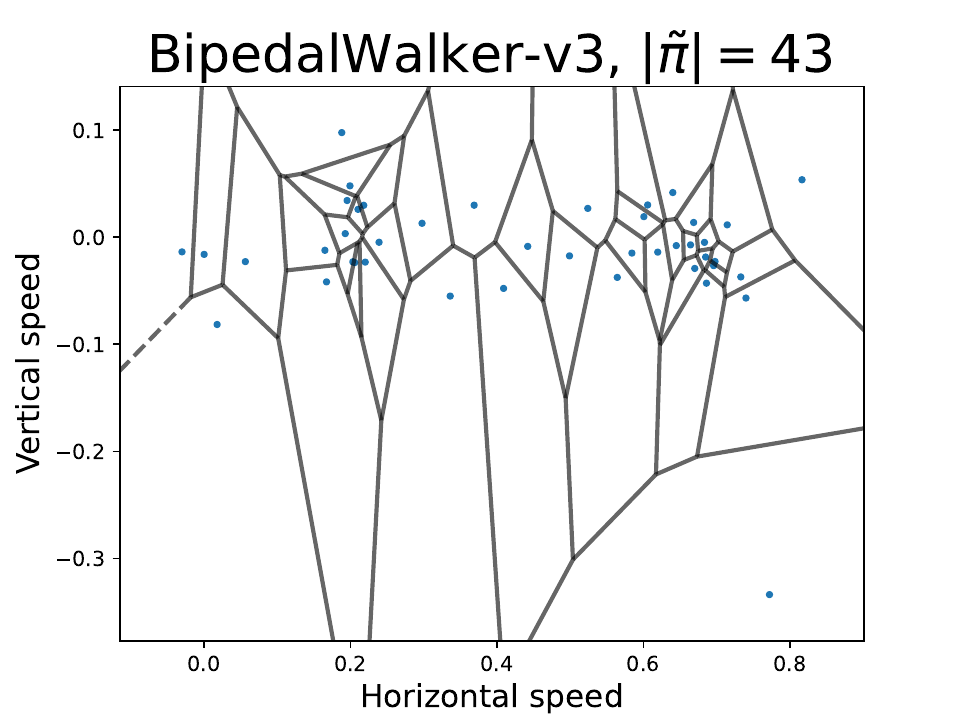}
		\label{fig:second}
	\end{subfigure}
	
	\caption{Partitioning of both LunarLander (on its x and y position) and BipedalWalker (vertical and horizontal speed). We can observe more and smaller regions in places such as around the and final height of the lunar vehicle and around two horizontal speeds for the walker. These indicate more non-linear behaviour in the original TD3 policy.}
	\label{fig:complex_voronoi}
\end{figure}

In this section, we have a look in the kind of insights we could have on the partitioning of the state space.
We first look how VSP-critic would partition the states of SimpleGoal. In Figure \ref{fig:simplegoal_quiver} we show both the TD3 policy and two sets of subpolicies learned with VSP-critic as quiver plots. The arrows indicate at each point what the direction is the policy steps towards and how far it does. The original policy has clearly learned to avoid the pitfall region by moving around it while having a tendency to move towards the goal when not obstructed. 
For a VSP-critic that learned with $\texttt{min\_codeword\_distance} = 0.3$ we see that it found 9 regions of linear behaviour. On first sight it seems that the regions are placed uniformly in the state space but on further inspection we could observe neighboring regions with their decision boundaries close to where we see sudden change in the original policy. This is most noticeable for regions rear the goal state.
In the case of $\texttt{min\_codeword\_distance} = 0.6$, the 3 regions share a shared vertex situated within the pitfall region. This causes an almost symmetric partitioning with the two upper regions behaving counterclockwise around the middle while the one below does it clockwise.\\

To visualize environments with a state space larger than two, a selection of variables to compare needs to be made. In Figure \ref{fig:complex_voronoi} we decided to look at the relation between the $x$ and $y$ position of the lunar vehicle where for the walker we compared both hull vertical and horizontal speed. Both diagrams come from the best performing runs and are taken at the algorithm iteration that yielded highest the highest validation return.

For both diagrams, the partitioning varies the size of the cells with distinct clusters of smaller ones around places of interest. This close proximity of codeword points indicate that this set of regions in the state space is of higher complexity and therefore needs a larger amount of linear functions.
For LunarLander, a collection can be found at $y$ position around 1.50, which is the height the lander starts descending from. Another collection is present around $y = 0$, which is ground level. This indicates that the lander has to perform more complex behavior around the landing site in order to avoid crashing.\\
For BipedalWalker, most points concentrate within the range $\left[ -0.1, 0.1\right] $. We see more codewords clustered around a horizontal speed of 0.2 and 0.7. However, in this case, it is more difficult to infer reasoning since there are 22 more variables that can influence the large present of points in these areas. We do however notice a subpolicy that entirely operates on negative vertical speeds and horizontal speeds above 0.4. 

\subsection{Found subpolicies}
\begin{figure}[]
	\centering
	\includegraphics[width=.7\textwidth]{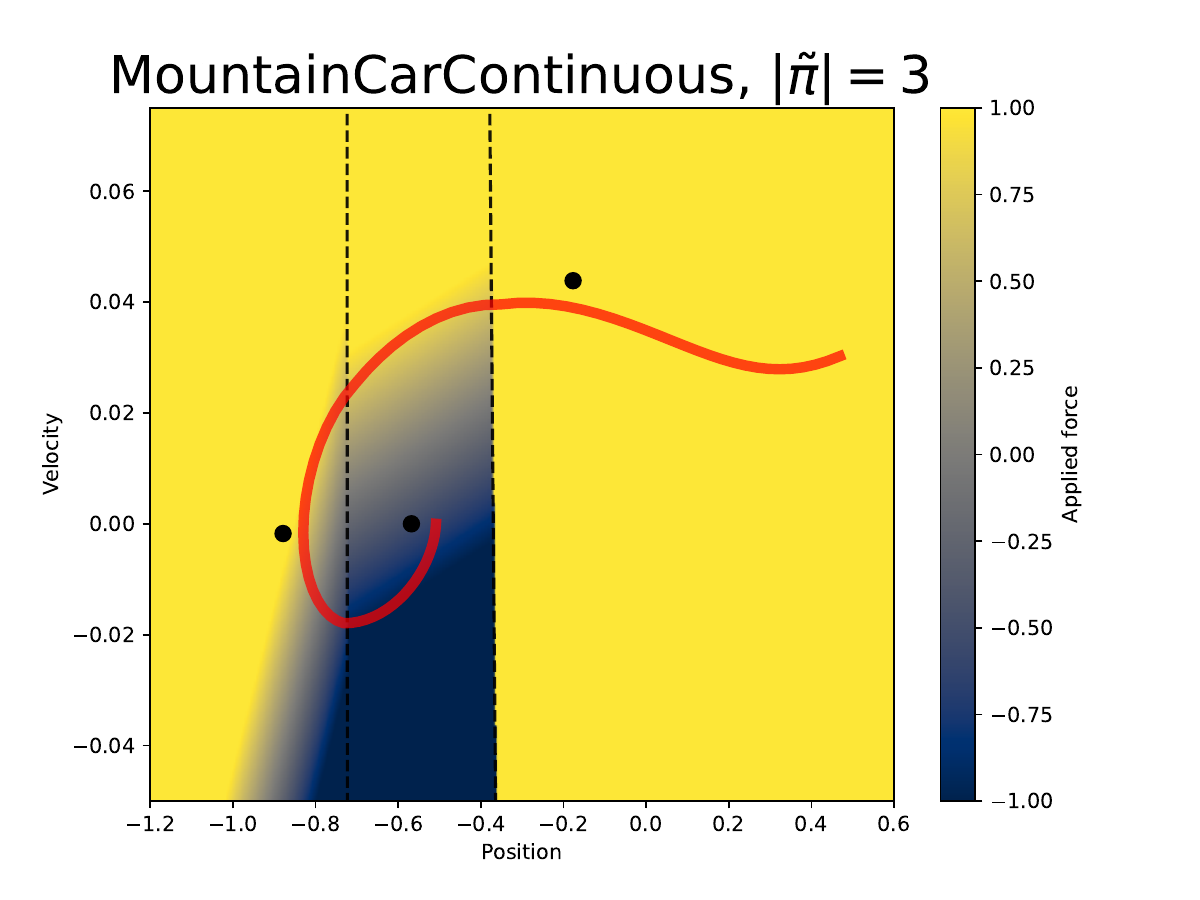}
	\caption{A set of 3 linear functions for MountainCarContinuous using $\texttt{min\_codeword\_distance} = 0.3$. Color indicates the amount of force applied to the car in newton. A sample trajectory (red line) shows how the car traverses trough the regions.}
	\label{fig:mountaincar_traject}
\end{figure}

\begin{table}
	\centering
	\begin{tabular}{|c|c|}
		\hline
		Codeword                    & $F$  		     \\						
		\hline
		\texttt{[-0.880, 0.002]}    & $-8.972x+30.034v-6.660$\\
		\texttt{[-0.573, 0.000]}    & $-1.846x+39.945v-1.567$\\
		\texttt{[-175, 0.044]}      & $1.000$\\
		\hline
	\end{tabular}
	\caption{Codewords and linked linear functions for the the partitioning in Figure \ref{fig:mountaincar_traject}}
\end{table}
\label{tab:mountaincar_functions}

To motivate the gained the insights that can be retrieved from a set of linear subpolicies, we examine a run of the VSP-critic algorithm on MountainCarContinuous. Using hyperparameters from Table \ref{tab:hyperparems} and $\texttt{min\_codeword\_distance} = 0.3$, the found set contains three policies. The regions are given in Table \ref{tab:mountaincar_functions} and visualized as a heatmap in Figure \ref{fig:mountaincar_traject} where yellow indicates a positive force and blue a negative one. A sample trajectory of one successful episode is given as a red line through the state space. \\

The three regions the VSP-critic algorithm found are almost perfect parallel next to each other. We start in the the region defined by codeword \texttt{[-0.573, 0.000]}. Here, the subpolicy outputs a negative force if the velocity $v$ of the cart is 0. Relative to all other subpolicies, a larger negative bias and negative weight for the position component $x$ pushes the car uphill to the left. Once the second region \texttt{[-0.573, 0.000]} is reached, the weight of $x$ is lower while the positive weight of the velocity increases. Around $x = -0.828$, the policy switches from a negative to a positive force as can be seen by the angle the linear policy makes. The car accelerates downhill and enters the previous region again around $x = -0.748$. Since the velocity of the cart is now above 0.02, the same policy that pushed the car with to the left now contributes to its acceleration to the right. This continuous at when the cart starts climbing at $x = -0.6$ until it reaches the final region at $x=-0.312$. The final region has near zero weights for both state variables and has a positive bias close to 1. This follows the intuition that at a certain point up the hill, the car only needs to apply positive force in order to get to the top of the hill as fast as possible.\\

The behaviour we just described is consist with control policies learned with DRL, where the cart has to climb the hill to the left first in order to gain enough momentum to accelerate to the top on the right. As we described the policy above, new scenarios can now be tested such as different starting conditions and the addition of disturbance to validate whether or not the learned policy fulfills the desiderata of the user.
\section{Conclusion}
In this paper, we proposed a novel approach to partition a black box policy from a deep reinforcement learning setting into a set of simpler models. We accomplish this by partitioning the state space using Voronoi quantization into distinct regions where these linear subpolices can operate sufficiently close to the original policy. We validated our approach on a set of well known continuous control problems and observed that these environments can be controlled by a set of linear functions.
We indicated that our approach is model-agnostic, only relying on a notion of model capacity by exploiting the critic from the original policy. However, we did not validate our approach other classes of models which is considered for future work. The argument that our method produces an interpretable partitioning depends on how much insights the learned models have, such as the weights in a linear function indicating feature importance. The decision boundaries around the partitions, formed by the the Voronoi quantizer, could be considered as complex for to give any insight. Instead, the demarcation is purely distance based (nearest codeword) in relation to all nearest neighboring regions. Future work would look into how to describe these decision boundaries.

\begin{credits}
	\subsubsection{\ackname} This research received funding from the Flanders Research Foundation via FWO S007723N (CTRLxAI) and FWO G062819N (Explainable Reinforcement Learning). We acknowledge financial support from the Flemish Government (AI Research Program).
	
	\subsubsection{\discintname} The authors of this dissemination declare to have no conflict of interest with any other party.
	
	\subsubsection{Reproduction} Code is made available for reproduction at \url{https://gitlab.ai.vub.ac.be/sdeproost/vsp.git} under MIT license.
	
\end{credits}
\vfill
\pagebreak
\printbibliography

@inproceedings{10.1609/aaai.v39i15.33733,
  title = {Understanding Individual Agent Importance in Multi-Agent System via Counterfactual Reasoning},
  booktitle = {Proceedings of the Thirty-Ninth {{AAAI}} Conference on Artificial Intelligence and Thirty-Seventh Conference on Innovative Applications of Artificial Intelligence and Fifteenth Symposium on Educational Advances in Artificial Intelligence},
  author = {Chen, Jianming and Wang, Yawen and Wang, Junjie and Xie, Xiaofei and Hu, Jun and Wang, Qing and Xu, Fanjiang},
  year = 2025,
  series = {{{AAAI}}'25/{{IAAI}}'25/{{EAAI}}'25},
  publisher = {AAAI Press},
  doi = {10.1609/aaai.v39i15.33733},
  articleno = {1760},
  isbn = {978-1-57735-897-8}
}

@inproceedings{akrour_etal_2018_RegularizingReinforcement,
  title = {Regularizing {{Reinforcement Learning}} with {{State Abstraction}}},
  booktitle = {2018 {{IEEE}}/{{RSJ International Conference}} on {{Intelligent Robots}} and {{Systems}} ({{IROS}})},
  author = {Akrour, Riad and Veiga, Filipe and Peters, Jan and Neumann, Gerhard},
  year = 2018,
  month = oct,
  pages = {534--539},
  issn = {2153-0866},
  doi = {10.1109/IROS.2018.8594201},
  urldate = {2026-02-02}
}

@article{barredoarrieta_etal_2020_ExplainableArtificial,
  title = {Explainable {{Artificial Intelligence}} ({{XAI}}): {{Concepts}}, Taxonomies, Opportunities and Challenges toward Responsible {{AI}}},
  shorttitle = {Explainable {{Artificial Intelligence}} ({{XAI}})},
  author = {Barredo Arrieta, Alejandro and {D{\'i}az-Rodr{\'i}guez}, Natalia and Del Ser, Javier and Bennetot, Adrien and Tabik, Siham and Barbado, Alberto and Garcia, Salvador and {Gil-Lopez}, Sergio and Molina, Daniel and Benjamins, Richard and Chatila, Raja and Herrera, Francisco},
  year = 2020,
  month = jun,
  journal = {Information Fusion},
  volume = {58},
  pages = {82--115},
  issn = {15662535},
  doi = {10.1016/j.inffus.2019.12.012},
  urldate = {2023-09-22},
  langid = {english}
}

@article{bekkemoen_2023_ExplainableReinforcement,
  title = {Explainable Reinforcement Learning ({{XRL}}): A Systematic Literature Review and Taxonomy},
  shorttitle = {Explainable Reinforcement Learning ({{XRL}})},
  author = {Bekkemoen, Yanzhe},
  year = 2023,
  month = nov,
  journal = {Machine Learning},
  issn = {0885-6125, 1573-0565},
  doi = {10.1007/s10994-023-06479-7},
  urldate = {2023-12-21},
  langid = {english}
}

@article{bellman_1957_MarkovianDecision,
  title = {A Markovian Decision Process},
  author = {Bellman, Richard},
  year = 1957,
  journal = {Indiana University Mathematics Journal},
  volume = {6},
  pages = {679--684}
}

@article{bentley_1975_MultidimensionalBinary,
  title = {Multidimensional Binary Search Trees Used for Associative Searching},
  author = {Bentley, Jon Louis},
  year = 1975,
  month = sep,
  journal = {Communications of the ACM},
  volume = {18},
  number = {9},
  pages = {509--517},
  issn = {0001-0782, 1557-7317},
  doi = {10.1145/361002.361007},
  urldate = {2026-02-24},
  langid = {english}
}

@inproceedings{coppens_etal_2019_DistillingDeep,
  title = {Distilling Deep Reinforcement Learning Policies in Soft Decision Trees},
  booktitle = {{{ProceedingsoftheIJCAI2019Workshop onExplainableArtificialIntelligence}}},
  author = {Coppens, Youri and Efthymiadis, Kyriakos and Lenaerts, Tom and Now{\'e}, Ann},
  year = 2019,
  pages = {1--6}
}

@article{dasgupta_etal_2015_PolicyTree,
  title = {Policy Tree: {{Adaptive}} Representation for Policy Gradient},
  author = {Das Gupta, Ujjwal and Talvitie, Erik and Bowling, Michael},
  year = 2015,
  month = feb,
  journal = {Proceedings of the AAAI Conference on Artificial Intelligence},
  volume = {29},
  number = {1},
  doi = {10.1609/aaai.v29i1.9613}
}

@inproceedings{erwig_etal_2018_ExplainingDeep,
  title = {Explaining Deep Adaptive Programs via Reward Decomposition},
  booktitle = {{{IJCAI}}/{{ECAI Workshop}} on {{Explainable Artificial Intelligence}}},
  author = {Erwig, Martin and Fern, Alan and Murali, Magesh and Koul, Anurag},
  year = 2018
}

@misc{frosst_hinton_2017_DistillingNeural,
  title = {Distilling a {{Neural Network Into}} a {{Soft Decision Tree}}},
  author = {Frosst, Nicholas and Hinton, Geoffrey},
  year = 2017,
  month = nov,
  number = {arXiv:1711.09784},
  eprint = {1711.09784},
  primaryclass = {cs, stat},
  publisher = {arXiv},
  urldate = {2024-03-06},
  archiveprefix = {arXiv},
  langid = {english}
}

@article{gjaerum_etal_2021_ExplainingDeep,
  title = {Explaining a {{Deep Reinforcement Learning Docking Agent Using Linear Model Trees}} with {{User Adapted Visualization}}},
  author = {Gj{\ae}rum, Vilde B. and Str{\"u}mke, Inga and Alsos, Ole Andreas and Lekkas, Anastasios M.},
  year = 2021,
  month = oct,
  journal = {Journal of Marine Science and Engineering},
  volume = {9},
  number = {11},
  eprint = {2203.00368},
  primaryclass = {cs},
  pages = {1178},
  issn = {2077-1312},
  doi = {10.3390/jmse9111178},
  urldate = {2026-02-13},
  archiveprefix = {arXiv}
}

@article{gray_1984_VectorQuantization,
  title = {Vector Quantization},
  author = {Gray, R.},
  year = 1984,
  month = apr,
  journal = {IEEE ASSP Magazine},
  volume = {1},
  number = {2},
  pages = {4--29},
  issn = {1558-1284},
  doi = {10.1109/MASSP.1984.1162229}
}

@inproceedings{green_sheppard_2024_PerformanceRobustness,
  title = {On the {{Performance}} and {{Robustness}} of {{Linear Model U-Trees}} in {{Mimic Learning}}},
  booktitle = {2024 {{International Conference}} on {{Machine Learning}} and {{Applications}} ({{ICMLA}})},
  author = {Green, Matthew and Sheppard, John W.},
  year = 2024,
  month = dec,
  pages = {152--159},
  publisher = {IEEE},
  address = {Miami, FL, USA},
  doi = {10.1109/ICMLA61862.2024.00027},
  urldate = {2025-06-30},
  copyright = {https://doi.org/10.15223/policy-029},
  isbn = {979-8-3503-7488-9},
  langid = {english}
}

@inproceedings{haarnoja_etal_2018_SoftActorcritic,
  title = {Soft Actor-Critic: {{Off-policy}} Maximum Entropy Deep Reinforcement Learning with a Stochastic Actor},
  booktitle = {Proceedings of the 35th International Conference on Machine Learning},
  author = {Haarnoja, Tuomas and Zhou, Aurick and Abbeel, Pieter and Levine, Sergey},
  editor = {Dy, Jennifer and Krause, Andreas},
  year = 2018,
  month = jul,
  series = {Proceedings of Machine Learning Research},
  volume = {80},
  pages = {1861--1870},
  publisher = {PMLR}
}

@inproceedings{hein_etal_2018_GeneratingInterpretable,
  title = {Generating Interpretable Fuzzy Controllers Using Particle Swarm Optimization and Genetic Programming},
  booktitle = {Proceedings of the {{Genetic}} and {{Evolutionary Computation Conference Companion}}},
  author = {Hein, Daniel and Udluft, Steffen and Runkler, Thomas A.},
  year = 2018,
  month = jul,
  pages = {1268--1275},
  publisher = {ACM},
  address = {Kyoto Japan},
  doi = {10.1145/3205651.3208277},
  urldate = {2025-12-16},
  isbn = {978-1-4503-5764-7},
  langid = {english}
}

@inproceedings{kakade_2001_NaturalPolicy,
  title = {A {{Natural Policy Gradient}}},
  booktitle = {Advances in {{Neural Information Processing Systems}}},
  author = {Kakade, Sham M},
  year = 2001,
  volume = {14},
  publisher = {MIT Press},
  urldate = {2026-02-24}
}

@inproceedings{kohler_etal_2024_InterpretableEditable,
  title = {Interpretable and {{Editable Programmatic Tree Policies}} for {{Reinforcement Learning}}},
  booktitle = {17th {{European Workshop}} on {{Reinforcement Learning}}},
  author = {Kohler, Hector and Delfosse, Quentin and Akrour, Riad and Kersting, Kristian and Preux, Philippe},
  year = 2024,
  month = oct,
  address = {Toulouse},
  langid = {english}
}

@inproceedings{konda_tsitsiklis_1999_ActorCriticAlgorithms,
  title = {Actor-{{Critic Algorithms}}},
  booktitle = {Advances in {{Neural Information Processing Systems}}},
  author = {Konda, Vijay and Tsitsiklis, John},
  year = 1999,
  volume = {12},
  publisher = {MIT Press},
  urldate = {2025-03-19}
}

@article{lee_lau_2004_AdaptiveStatea,
  title = {Adaptive State Space Partitioning for Reinforcement Learning},
  author = {Lee, Ivan S.K. and Lau, Henry Y.K.},
  year = 2004,
  month = sep,
  journal = {Engineering Applications of Artificial Intelligence},
  volume = {17},
  number = {6},
  pages = {577--588},
  issn = {09521976},
  doi = {10.1016/j.engappai.2004.08.005},
  urldate = {2025-06-04},
  copyright = {https://www.elsevier.com/tdm/userlicense/1.0/},
  langid = {english}
}

@misc{lillicrap_etal_2015_ContinuousControla,
  title = {Continuous Control with Deep Reinforcement Learning},
  author = {Lillicrap, Timothy P. and Hunt, Jonathan J. and Pritzel, Alexander and Heess, Nicolas and Erez, Tom and Tassa, Yuval and Silver, David and Wierstra, Daan},
  year = 2015,
  month = sep,
  journal = {arXiv.org},
  urldate = {2026-05-14},
  howpublished = {https://arxiv.org/abs/1509.02971v6},
  langid = {english}
}

@incollection{liu_etal_2019_InterpretableDeep,
  title = {Toward {{Interpretable Deep Reinforcement Learning}} with {{Linear Model U-Trees}}},
  booktitle = {Machine {{Learning}} and {{Knowledge Discovery}} in {{Databases}}},
  author = {Liu, Guiliang and Schulte, Oliver and Zhu, Wang and Li, Qingcan},
  editor = {Berlingerio, Michele and Bonchi, Francesco and G{\"a}rtner, Thomas and Hurley, Neil and Ifrim, Georgiana},
  year = 2019,
  volume = {11052},
  pages = {414--429},
  publisher = {Springer International Publishing},
  address = {Cham},
  doi = {10.1007/978-3-030-10928-8_25},
  urldate = {2025-06-30},
  isbn = {978-3-030-10927-1 978-3-030-10928-8},
  langid = {english}
}

@inproceedings{macqueen_1967_MethodsClassification,
  title = {Some Methods for Classification and Analysis of {{MultiVariate}} Observations},
  booktitle = {Proc. of the Fifth Berkeley Symposium on Mathematical Statistics and Probability},
  author = {MacQueen, J. B.},
  editor = {Cam, L. M. Le and Neyman, J.},
  year = 1967,
  volume = {1},
  pages = {281--297},
  publisher = {University of California Press}
}

@article{miller_2019_ExplanationArtificial,
  title = {Explanation in Artificial Intelligence: {{Insights}} from the Social Sciences},
  shorttitle = {Explanation in Artificial Intelligence},
  author = {Miller, Tim},
  year = 2019,
  month = feb,
  journal = {Artificial Intelligence},
  volume = {267},
  pages = {1--38},
  issn = {0004-3702},
  doi = {10.1016/j.artint.2018.07.007},
  urldate = {2023-09-22}
}

@article{Mitsopoulos2021TowardAP,
  title = {Toward a Psychology of Deep Reinforcement Learning Agents Using a Cognitive Architecture},
  author = {Mitsopoulos, Konstantinos and Somers, Sterling and Schooler, Joel N. and Lebiere, Christian and Pirolli, Peter and Thomson, Robert},
  year = 2021,
  journal = {Topics in cognitive science}
}

@article{mnih_etal_2015_HumanlevelControla,
  title = {Human-Level Control through Deep Reinforcement Learning},
  author = {Mnih, Volodymyr and Kavukcuoglu, Koray and Silver, David and Rusu, Andrei A. and Veness, Joel and Bellemare, Marc G. and Graves, Alex and Riedmiller, Martin and Fidjeland, Andreas K. and Ostrovski, Georg and Petersen, Stig and Beattie, Charles and Sadik, Amir and Antonoglou, Ioannis and King, Helen and Kumaran, Dharshan and Wierstra, Daan and Legg, Shane and Hassabis, Demis},
  year = 2015,
  month = feb,
  journal = {Nature},
  volume = {518},
  number = {7540},
  pages = {529--533},
  issn = {0028-0836, 1476-4687},
  doi = {10.1038/nature14236},
  urldate = {2023-09-19},
  langid = {english}
}

@article{padakandla_2022_SurveyReinforcement,
  title = {A {{Survey}} of {{Reinforcement Learning Algorithms}} for {{Dynamically Varying Environments}}},
  author = {Padakandla, Sindhu},
  year = 2022,
  month = jul,
  journal = {ACM Computing Surveys},
  volume = {54},
  number = {6},
  pages = {1--25},
  issn = {0360-0300, 1557-7341},
  doi = {10.1145/3459991},
  urldate = {2026-02-24},
  langid = {english}
}

@article{recht_2019_TourReinforcement,
  title = {A {{Tour}} of {{Reinforcement Learning}}: {{The View}} from {{Continuous Control}}},
  shorttitle = {A {{Tour}} of {{Reinforcement Learning}}},
  author = {Recht, Benjamin},
  year = 2019,
  month = may,
  journal = {Annual Review of Control, Robotics, and Autonomous Systems},
  volume = {2},
  number = {1},
  pages = {253--279},
  issn = {2573-5144, 2573-5144},
  doi = {10.1146/annurev-control-053018-023825},
  urldate = {2026-02-24},
  langid = {english}
}

@inproceedings{ribeiro_etal_2016_WhyShould,
  title = {"{{Why}} Should {{I}} Trust You?": {{Explaining}} the Predictions of Any Classifier},
  booktitle = {Proceedings of the 22nd {{ACM SIGKDD}} International Conference on Knowledge Discovery and Data Mining},
  author = {Ribeiro, Marco Tulio and Singh, Sameer and Guestrin, Carlos},
  year = 2016,
  series = {Kdd '16},
  pages = {1135--1144},
  publisher = {Association for Computing Machinery},
  address = {San Francisco, California, USA and New York, NY, USA},
  doi = {10.1145/2939672.2939778},
  isbn = {978-1-4503-4232-2}
}

@inproceedings{rupprecht_etal_2020_FindingVisualizing,
  title = {Finding and {{Visualizing Weaknesses}} of {{Deep Reinforcement Learning Agents}}},
  shorttitle = {{{ICLR}}},
  booktitle = {International {{Conference}} on {{Learning Representations}}},
  author = {Rupprecht, Christian and Ibrahim, Cyril and Pal, Christopher J},
  year = 2020,
  langid = {english}
}

@inproceedings{rusu_etal_2015_PolicyDistillation,
  title = {Policy {{Distillation}}},
  shorttitle = {{{ICLR}}},
  booktitle = {International {{Conference}} on {{Learning Representations}}},
  author = {Rusu, Andrei A. and Colmenarejo, Sergio Gomez and G{\"u}l{\c c}ehre, {\c C}aglar and Desjardins, Guillaume and Kirkpatrick, James and Pascanu, Razvan and Mnih, Volodymyr and Kavukcuoglu, Koray and Hadsell, Raia},
  year = 2015,
  volume = {abs/1511.06295},
  address = {San Diego, CA, USA}
}

@misc{schulman_etal_2017_ProximalPolicy,
  title = {Proximal {{Policy Optimization Algorithms}}},
  author = {Schulman, John and Wolski, Filip and Dhariwal, Prafulla and Radford, Alec and Klimov, Oleg},
  year = 2017,
  month = aug,
  number = {arXiv:1707.06347},
  eprint = {1707.06347},
  primaryclass = {cs},
  publisher = {arXiv},
  doi = {10.48550/arXiv.1707.06347},
  urldate = {2026-02-24},
  archiveprefix = {arXiv}
}

@book{sutton_barto_2014_ReinforcementLearning,
  title = {Reinforcement Learning: {{An}} Introduction},
  shorttitle = {Reinforcement Learning},
  author = {Sutton, Richard S. and Barto, Andrew},
  year = 2014,
  series = {Adaptive Computation and Machine Learning},
  edition = {Nachdruck},
  publisher = {The MIT Press},
  address = {Cambridge, Massachusetts},
  isbn = {978-0-262-19398-6},
  langid = {english}
}

@misc{towers_etal_2024_GymnasiumStandard,
  title = {Gymnasium: {{A Standard Interface}} for {{Reinforcement Learning Environments}}},
  shorttitle = {Gymnasium},
  author = {Towers, Mark and Kwiatkowski, Ariel and Terry, Jordan and Balis, John U. and Cola, Gianluca De and Deleu, Tristan and Goul{\~a}o, Manuel and Kallinteris, Andreas and Krimmel, Markus and KG, Arjun and {Perez-Vicente}, Rodrigo and Pierr{\'e}, Andrea and Schulhoff, Sander and Tai, Jun Jet and Tan, Hannah and Younis, Omar G.},
  year = 2024,
  month = nov,
  number = {arXiv:2407.17032},
  eprint = {2407.17032},
  primaryclass = {cs},
  publisher = {arXiv},
  doi = {10.48550/arXiv.2407.17032},
  urldate = {2025-08-13},
  archiveprefix = {arXiv}
}

@inproceedings{verma_etal_2019_ImitationProjectedProgrammatic,
  title = {Imitation-{{Projected Programmatic Reinforcement Learning}}},
  booktitle = {Advances in {{Neural Information Processing Systems}}},
  author = {Verma, Abhinav and Le, Hoang and Yue, Yisong and Chaudhuri, Swarat},
  year = 2019,
  volume = {32},
  publisher = {Curran Associates, Inc.},
  urldate = {2024-05-28}
}

@article{watkins_dayan_1992_Qlearning,
  title = {Q-Learning},
  author = {Watkins, Christopher J. C. H. and Dayan, Peter},
  year = 1992,
  month = may,
  journal = {Machine Learning},
  volume = {8},
  number = {3},
  pages = {279--292},
  issn = {1573-0565},
  doi = {10.1007/BF00992698}
}

\vfill
\pagebreak

\appendix
\end{document}